\pdfoutput=1

\documentclass[11pt]{article}

\usepackage[final]{acl}

\usepackage{times}
\usepackage{latexsym}

\usepackage[T1]{fontenc}

\usepackage[utf8]{inputenc}

\usepackage{microtype}

\usepackage{inconsolata}

\usepackage{graphicx}

\usepackage{amsmath}
\usepackage{amssymb}
\usepackage{mathtools}
\usepackage{amsthm}
\usepackage{algorithm}
\usepackage{algpseudocode}
\usepackage{color}
\usepackage{xcolor}
\usepackage{wrapfig}
\usepackage{colortbl}
\usepackage{multirow}
\usepackage{booktabs}
\usepackage{rotating} 
\usepackage{subfigure}
\usepackage{enumitem}
\usepackage{soul}  

\definecolor{codeblue}{RGB}{0, 82, 147}     
\definecolor{codegreen}{RGB}{0, 128, 0}    
\definecolor{codegray}{RGB}{100, 100, 100}  
\definecolor{codeorange}{RGB}{230, 145, 56} 
\definecolor{darkerblue}{rgb}{0,0.08,0.45} 
\definecolor{royalblue}{RGB}{65,105,225}
\definecolor{lightblue}{RGB}{221,235,247}
\definecolor{fig3blue}{RGB}{47, 122, 232}  
\definecolor{fig3red}{RGB}{213, 32, 52}
\definecolor{fig3green}{RGB}{0, 137, 72} 
\definecolor{fig3yellow}{RGB}{217, 161, 5}
\definecolor{gray94}{gray}{.94}
\definecolor{gray90}{gray}{.90}

\newcommand{\red}[1]{\textcolor{red}{#1}}

\definecolor{darkgreen}{RGB}{34,139,34}
\newcommand{\green}[1]{\textcolor{darkgreen}{#1}}
\newcommand{\gray}[1]{\textcolor{gray}{#1}}
\newcommand{\gbf}[1]{\green{\bf{#1}}}
\newcommand{\rbf}[1]{\red{\bf{#1}}}

\newcolumntype{g}{>{\columncolor{gray94}}c} 
\newcolumntype{b}{>{\columncolor{lightblue}}c} 
\newcommand{\grow}[1]{\rowcolor{gray94}{#1}} 
\newcommand{\brow}[1]{\rowcolor{lightblue}{#1}} 
\newcommand{\gcell}[1]{\cellcolor{gray94}{#1}} 
\newcommand{\bcell}[1]{\cellcolor{lightblue}{#1}} 
\usepackage{pifont}
\newcommand{\cmark}{\ding{51}}%
\newcommand{\xmarkg}{\textcolor{gray}{\ding{55}}}%

\title{Taming LLMs by Scaling Learning Rates with Gradient Grouping
\thanks{$^\star$\textbf{Equal contribution}}
\thanks{$^\dagger$\textbf{Corresponding author:} Zicheng Liu
}
}

\author{
 \textbf{Siyuan Li\textsuperscript{1,2$\star$}},~
 \textbf{Juanxi Tian\textsuperscript{2,4$\star$}},~
 \textbf{Zedong Wang\textsuperscript{3$\star$}},~
 \textbf{Xin Jin\textsuperscript{2}},
\\
 \textbf{Zicheng Liu\textsuperscript{1,2$\dag$}},~
 \textbf{Wentao Zhang\textsuperscript{4}},~
 \textbf{Dan Xu\textsuperscript{3}}
\\
 \textsuperscript{1}Zhejiang University,~~
 \textsuperscript{2}Westlake University,
\\
 \textsuperscript{3}The Hong Kong University of Science and Technology,~~
 \textsuperscript{4}Peking University
\\
}

\makeatletter
\def\thanks#1{\protected@xdef\@thanks{\@thanks
        \protect\footnotetext{#1}}}
\makeatother

\begin{document}
\maketitle

\begin{abstract}

Training large language models (LLMs) poses challenges due to their massive scale and heterogeneous architectures. While adaptive optimizers like AdamW help address gradient variations, they still struggle with efficient and effective parameter-wise learning rate estimation, resulting in training instability, slow convergence, and poor compatibility with parameter-efficient fine-tuning (PEFT) techniques. This work introduces \textbf{\underline{S}}caling with \textbf{\underline{G}}radient \textbf{\underline{G}}rouping (\textbf{SGG}), an optimizer wrapper that improves adaptive learning rate estimation by dynamic grouping and group-specific scaling. SGG first groups gradient statistics in each layer into clusters and then applies cluster-specific scaling to calibrate learning rates for each parameter, thus imposing collective group-wise constraints while maintaining precise per-parameter adaptation. Experiments on diverse (M)LLM benchmarks show that SGG integrates seamlessly with existing optimizers, and offers consistent gains and faster convergence over baselines, with various model sizes. Its stability across varying batch sizes and learning rates establishes SGG as a robust choice for LLM optimization.

\end{abstract}

\section{Introduction}
\label{sec:intro}
Optimization algorithms have long been the cornerstone of deep learning systems. Among these, adaptive optimizers~\citep{iclr2015adam, iclr2019AdamW, iclr2020lamb} stand out for their ability to adjust individual learning rates for each parameter, enabling effective training of large language models (LLMs) with heterogeneous architecture~\citep{EMNLP2020GradInit, zhang2025why_adam}. Yet, their reliance on per-parameter statistics (\textit{e.g.}, first and second moments of gradient) incurs substantial memory overhead, which limits their application, especially in resource-constrained scenarios.

\begin{figure}[t!]
    \vspace{-0.5em}
    \centering
    \includegraphics[width=0.99\linewidth]{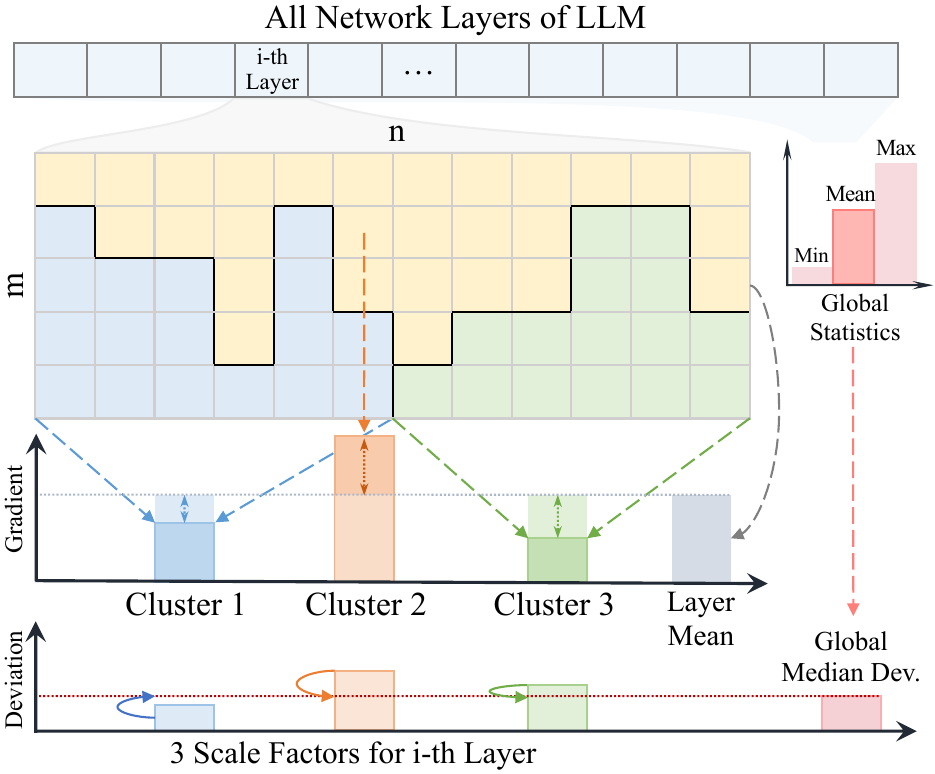}
    \vspace{-1.75em}
    \caption{\textbf{Scaling with Gradient Grouping.} Illustration of SGG with online grouping and group-specific learning rate (LR) scaling upon adaptive LR optimizers.
    }
    \label{fig:intro}
    \vspace{-1.55em}
\end{figure}

To address this, parameter-efficient fine-tuning (PEFT)~\citep{hu2021lora, dettmers2024qlora} has garnered increasing attention, which reduces trainable parameters via low-rank updates. While memory-efficient, PEFT incurs performance degradation compared to full-rank training (Table~\ref{tab:comp_c4_pt}) and requires architecture modification. In parallel, efforts have been made to compress optimizer states directly, \textit{e.g.}, by low-bit quantization or approximating gradient statistics~\citep{icml2018adafactor, luo2023came, zhu2024apollo}. However, these generic approaches typically rely on heuristic priors that might discard crucial information, resulting in inconsistent efficacy across tasks (Table~\ref{table:comp_GLUE_full}). This leaves practitioners at a deadlock: \textit{the compromise of performance in LLM training seems inevitable.}

\begin{figure*}[t]
    \vspace{-0.5em}
    \centering
    \subfigtopskip=-0.5pt
    \subfigbottomskip=-0.5pt
    \subfigcapskip=-4.0pt
    \subfigure[Gradiant Distribution]{\hspace{-0.5em}
    \label{fig:cluster(a)}\includegraphics[width=0.33\linewidth,trim= 0 0 0 0,clip]{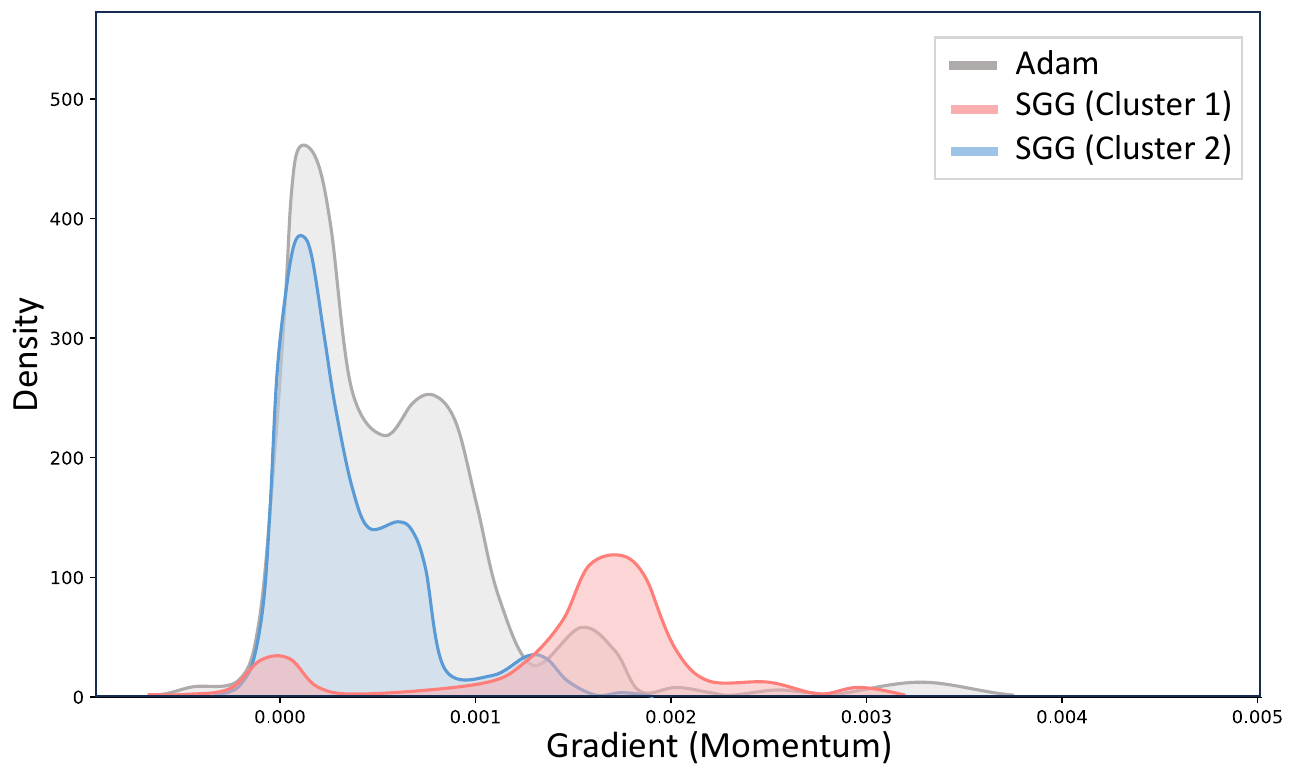}}
    \subfigure[Learning Rate Distribution]{\hspace{-0.5em}
    \label{fig:cluster(b)}\includegraphics[width=0.33\linewidth,trim= 0 0 0 0,clip]{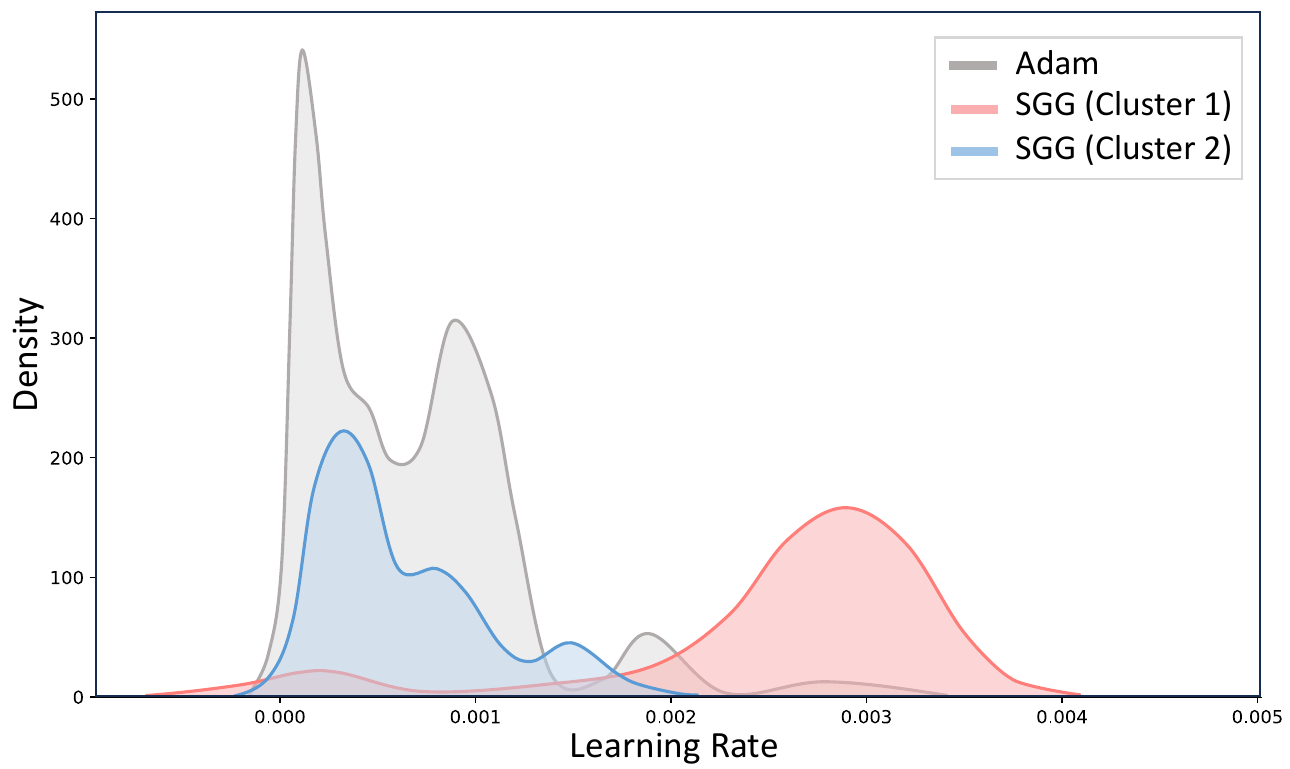}}
    \subfigure[Grad Norm Distribution]{\hspace{-0.5em}
    \label{fig:cluster(c)}\includegraphics[width=0.33\linewidth,trim= 0 0 0 0,clip]{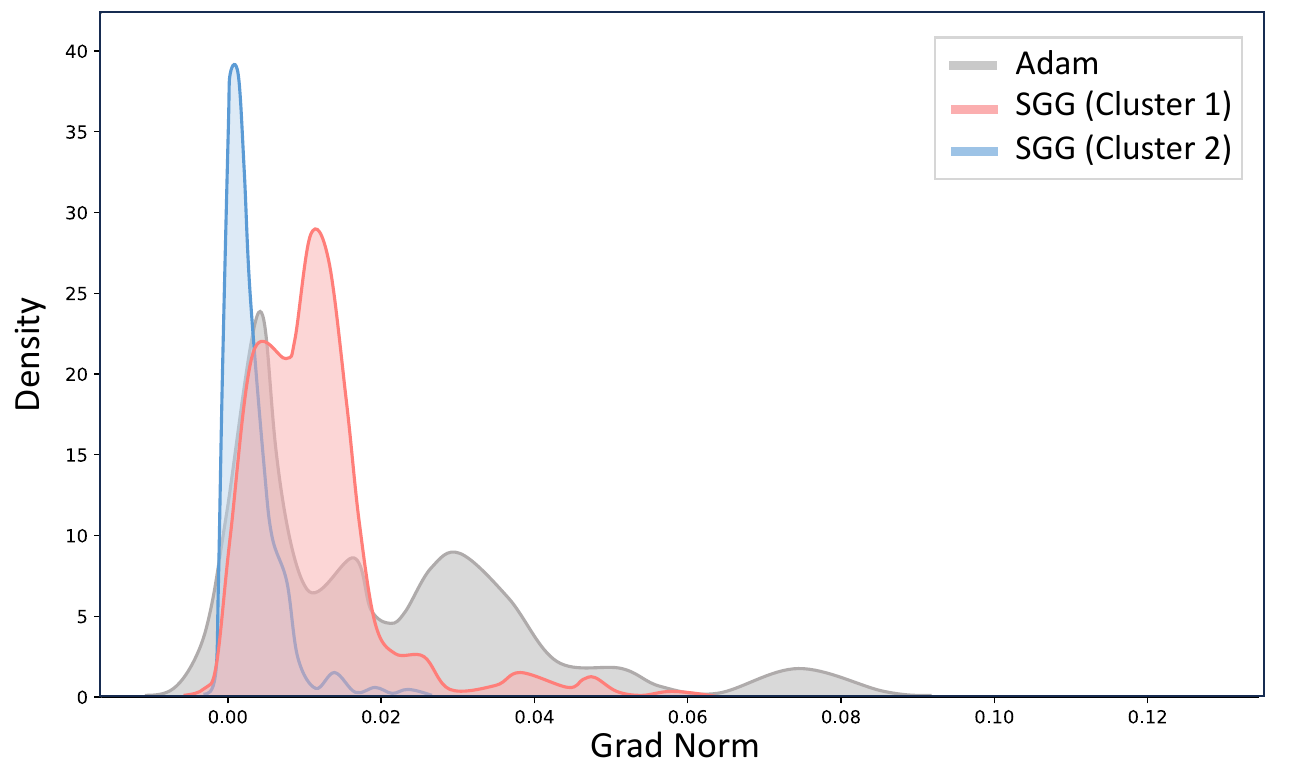}}
    \vspace{-0.5em}
    \caption{\textbf{Clusters of gradient statistics with LLaMA-1B pre-training on C4.} Distributions of \textbf{(a)} parameter-wise gradients $g^t$ and \textbf{(b)} learning rates $\alpha^t$ for $12$-th FFN layer at $5k$ iterations. SGG identifies diverse clusters compared to Adam (in \sethlcolor{gray90}\hl{gray}), introducing group constraints while maintaining parameter-wise adaptation. \textbf{(c)} Distribution of gradient $L_2$-norms across layers, showcasing SGG's ability to adapt to LLMs' heterogeneity~\citep{zhang2025why_adam}.
    }
    \label{fig:sgg_clustering}
    \vspace{-0.65em}
\end{figure*}

Recent studies light the way by revealing that different layers in LLMs (\textit{e.g.}, attention and MLP) exhibit distinct yet internally consistent optimization behaviors~\citep{2024BOCB, zhang2025why_adam}, suggesting potential redundancy in existing adaptive methods. Adam-mini~\citep{zhang2024adam_mini} thus divides model parameters into pre-defined groups -- each assigned an average learning rate instead, with minimal performance drop versus prior works.

In this work, we propose to \textit{scale learning rates with grouping constraints} rather than replace them. We first conduct pilot studies on LLaMA-1B pre-training to gain intuition. As shown in Figure~\ref{fig:sgg_clustering}, distinct clustering patterns are observed in layer-wise gradient statistics, which aligns with previous findings. However, they also exhibit noticeable characteristics (\textit{e.g.}, significant parameter-wise variations within each cluster, Sec.~\ref{sec:clustering}). This suggests that, while grouping is viable, retaining parameter-wise adaptation could still be beneficial for effective LLM training, especially in terms of performance, instead of replacing it with a single learning rate per group. Thus, we introduce \textbf{\underline{S}}caling with \textbf{\underline{G}}radient \textbf{\underline{G}}rouping (\textbf{SGG}), an optimizer wrapper to bridge per-parameter and group-wise learning rate control.
As illustrated in Figure~\ref{fig:intro} and Algorithm~\ref{alg:sgg}, SGG dynamically clusters gradient statistics (specifically momentum vectors $m_l$) in each layer $l$ (Sec.~\ref{sec:clustering}) and then performs cluster-specific scaling according to their deviation relative to that layer's and the entire model's global statistics (Sec.~\ref{sec:scaling}), which thus imposes grouping constraints for homogenization while maintaining parameter-wise adaptability.

Experiments demonstrate that SGG consistently delivers performance gains and accelerates convergence across different LLM (Sec.~\ref{sec:comp_llm}) and MLLM (Sec.~\ref{sec:comp_mllm}) benchmarks, such as pre-training on C4, supervised fine-tuning (SFT) on GLUE, PEFT on commonsense reasoning tasks, and Direct Preference Optimization (DPO). For instance, Adam combined with SGG could surpass recent optimizers on C4 pre-training across diverse model sizes (from 60M to 1B). More importantly, SGG enables low-rank pre-training to match full-rank performance without modifications to the training pipeline, yielding up to $30.4\%$ lower validation perplexity over LoRA baselines -- a huge step forward as previous low-rank optimizers often struggled.

Our contributions are as follows:
\begin{itemize}[leftmargin=1.5em]
    \vspace{-0.25em}
    \item We present SGG, a flexible optimizer wrapper that scales adaptive learning rates with online grouping constraints rather than replace them in pre-defined groups, balancing parameter-wise dynamics and collective optimization behavior.
    \vspace{-1.7em}
    \item In practice, SGG integrates seamlessly with existing optimizers and PEFT techniques, requiring no changes to the training pipeline or model architectures. We also provide CPU, GPU, and hybrid implementations for different demands.
    \vspace{-1.7em}
    \item SGG's consistent effectiveness shows the potential of scaling adaptive learning rates with group-wise constraints. While SGG offers an intuitive instantiation of this scheme, different grouping and scaling strategies are conceivable and might inspire future studies along this line.
\end{itemize}

\section{Methodology}
\label{sec:method}

\begin{table*}[t]
    \centering
    \vspace{-0.75em}
    \caption{\textbf{Overview of typical optimizers (Opt.), PEFT techniques, and plug-and-play optimizer wrapper.} We consider a neural net layer $W \in \mathbb{R}^{m\times n}$ $(m\le n)$ with LoRA rank $r\ll m$ and SGG clusters $K\ll m$. Both weights and optimizer states are included. \textbf{(i)} Optimization states. We compare the adaptive Learning Rate (LR) costs, \textit{i.e.}, the extra state for its estimation (\textit{e.g.}, second moment, Non-negative Matrix Factorization (NMF), and SGG's cluster indices). \textbf{(ii)} Different low-rank integration. \textbf{(iii)} Performance. We report the averaged PPL (\%)$\downarrow$ for C4 pre-training in Table~\ref{tab:comp_c4_pt} as the illustration of performance gains with the relative GPU memory to Adam (full-rank) in PyTorch.
    }
    \vspace{-0.60em}
    \setlength{\tabcolsep}{0.6mm}
\resizebox{1.0\linewidth}{!}{
    \begin{tabular}{lc|ccc|ccc|ccc}
    \toprule
Category                & Method   & Adaptive LR             & Basic State             & Extra State                  & Low-Rank   & Plugin  & Extra Branch & C4$\downarrow$ & GPU Memory               \\ \hline
Classical Opt.          & SGD      & \xmarkg                 & \small{Weight \& Grad.} & \xmarkg                      & \xmarkg    & \xmarkg & \xmarkg      & $-$            & $2mn$                    \\
\grow{Adaptive LR Opt.} & Adam     & \small{Param-wise} $mn$ & \small{Weight \& Grad.} & \small{2$^{\text{nd}}$-Moment}~$mn$ & \xmarkg & \xmarkg & \xmarkg  & 23.36          & \gcell{$3mn$}            \\
Efficient Opt.          & CAME     & \small{Param-wise} $mn$ & \small{Weight \& Grad.} & \small{NMF}~$2(m$+$n)$       & NMF        & \xmarkg & \xmarkg      & \gbf{-1.64}    & $2mn$+$2(m$+$n)$         \\ \hline
PEFT                    & LoRA     & \xmarkg                 & \small{Full-rank Grad.} & \xmarkg                      & LoRA       & \cmark  & $r(m$+$n)$   & \textcolor{purple}{+5.06} & \textcolor{purple}{$+3r(m+n)$} \\
\brow{Opt. Wrapper}     & \bf{SGG} & \small{Group-wise} $K$  & Base Opt.               & \small{Indices}~$(mn$+$K)$   & Clustering & \cmark  & \xmarkg      & \gbf{-1.99}    & \gray{$+0$}              \\
    \bottomrule
    \end{tabular}
    }
    \label{tab:intro}
    \vspace{-0.5em}
\end{table*}

\subsection{Preliminaries and Problem Definition}
To demonstrate the plug-and-play integration of our SGG, we first outline the essential steps in gradient-based optimizers, marked in \textcolor{codeblue}{\textbf{blue}} in Algorithm~\ref{alg:sgg}.

The process typically begins with gradient computation. At iteration $t$, the gradient $g^{t}_l$ of objective $\mathcal{L}$ w.r.t. parameters $\theta^{t-1}_l$ of layer $l$ is calculated as:
\begin{equation}
g_l^{t} = \nabla_{\theta^{t-1}_l} \mathcal{L}(\theta^{t-1}_l, \mathcal{D})
\end{equation}
where $\mathcal{D}$ denotes the training dataset, and \( \theta^{t-1}_l \) is from previous iteration. Subsequently, historical gradient information is incorporated to stabilize the update, commonly referred to as momentum $m^{t}_l$. While vanilla SGD~\citep{tsmc1971sgd} uses the current gradient instead ($m^t_l = g^t_l$), momentum-based methods often employ an exponential moving average (EMA) to smooth estimates over time:
\begin{equation}
\hspace{-1.0em}
    m^{t}_l = \text{MomentumEstimate}(g_l^{t}, m_l^{t-1}, \beta_1)
\end{equation}
where $m_l^{t-1}$ is from the last iteration, and the EMA decay $\beta_1$ controls the retention of past gradients. 

\begin{algorithm}[t]
\caption{
    \normalsize{Scaling with Gradient Grouping}
}
\label{alg:sgg}
\begin{algorithmic}[1]
\footnotesize  
\Require{Parameters $\{\theta_l\}_{l=1}^L$, global learning rate schedule $\eta$, optimizer hyperparameters $(\beta_1, \beta_2)$, objective $\mathcal{L}$, dataset $\mathcal{D}$. SGG hyperparameters: cluster number $K$, recluster interval $T$, scaling EMA decay $\beta_3$.}
\Ensure{Optimized model parameters $\theta$.}

\State \textbf{Initialize:}
\State $\texttt{RandomInit}(\{\theta_l^0\}_{l=1}^L)$
\Comment{\textcolor{codegray}{Model parameters}}
\State $\{\alpha_l^0\}_{l=1}^L \leftarrow \eta$ \Comment{\textcolor{codegray}{Adaptive learning rates}}
\State $\{\mathcal{C}_l\}_{l=1}^L \leftarrow \mathbf{0}$ \Comment{\textcolor{codegray}{Cluster assignment}}
\State $\{\mathcal{S}_l\}_{l=1}^L \leftarrow \mathbf{1}$
\Comment{\textcolor{codegray}{Cluster scaling factor}}

\For{each iteration $t = 1, 2, \ldots $}
    \State $\eta^{t} \leftarrow$ LRScheduler($\eta$, $t$)
    \For{each layer $l = 1, 2, \ldots, L$}
        \State \textcolor{codeblue}{\textit{// --- Standard Gradient-based Update Steps ---}}
        \State \textbf{\textcolor{codeblue}{Gradient Computation}}
        \State $g_l^{t} \leftarrow \nabla_{\theta_l^{t-1}} \mathcal{L}(\theta^{t-1}, \mathcal{D})$
        
        \State \textbf{\textcolor{codeblue}{Momentum Estimation}}
        \State $m_l^t \leftarrow \text{MomentumEstimate}(g_l^t, m_l^{t-1}, \beta_1)$
        
        \State \textbf{\textcolor{codeblue}{Adaptive Learning Rate Estimation}}
        \State $\alpha_l^t \leftarrow \text{LREstimate}(\alpha_l^{t-1}, m_l^t, \beta_2, \eta^{t})$

        \State \textcolor{codegreen}{\textit{// --- SGG Specific Steps ---}}
        \If{$t \mod T == 0$} \Comment{\textcolor{codegray}{Re-clutering}}
            \State \textbf{\textcolor{codegreen}{Assign Gradient Clusters}}
            \State $\mathcal{C}_l^{t} \leftarrow \text{GradCluster}(m_l^{t}, K)$
            
            \State \textbf{\textcolor{codegreen}{Update Cluster Scaling Factors}}
            \State $\mathcal{S}_l^{t} \leftarrow \text{ScaleUpdate}(\mathcal{C}_l^{t}, m_l^{t}, \beta_3)$ 
        \EndIf
        
        \State \textbf{\textcolor{codegreen}{Apply Learning Rate Scaling}}
        \State $\alpha_l^t \leftarrow \alpha_l^t \cdot \mathcal{S}_l^{t}[\mathcal{C}_l^{t}]$ \Comment{\textcolor{codegray}{Cluster-specific scaling}}
        
        \State \textbf{\textcolor{codeblue}{Parameter Update}}
        \State $\theta_l^t \leftarrow \theta_l^{t-1} - \alpha_l^t \cdot m_l^{t}$
    \EndFor
\EndFor
\end{algorithmic}
\vspace{-0.25em}
\end{algorithm}

Adaptive learning rate algorithms (\textit{e.g.}, Adam and AdaGrad) further refine the process by calculating parameter-wise or layer-wise second-moment estimates of gradients to calibrate step sizes:
\begin{equation}
\alpha^{t}_l = \text{LREstimate}(\alpha_l^{t-1}, m_l^{t}, \beta_2, \eta^{t})
\end{equation}
where $\eta_t$ indicates the global learning rate set by scheduler at iteration $t$, and $\beta_2$ is the EMA decay like $\beta_1$. Non-adaptive methods, in contrast, simply use the global one instead ($\alpha^{t}_l = \eta^{t}$).
Note that this learning rate adaptation typically increases memory overhead, particularly for large-scale models -- a key challenge that most prior works aim to address.

In the last step, model parameters $\theta_l$ are updated by the learning rate scaled momentum ($\alpha_l^{t} \cdot m_l^{t}$): 
\begin{equation}
\theta_l^{t} = \theta_l^{t-1} - \alpha_l^{t} \cdot m_l^{t}
\end{equation}
This paradigm is common in existing optimizers, mostly differing in how $m^t_l$ and $\alpha^t_l$ are derived. Notably, SGG (highlighted in \textcolor{codegreen}{\textbf{green}} in Algorithm~\ref{alg:sgg}) builds upon this by leveraging these pre-computed states from the base optimizer to impose grouping constraints on parameter-wise $\alpha_l^{t}$, ensuring effortless integration with diverse optimizers, from SGD to APOLLO~\citep{zhu2024apollo}. In the following sections, we discuss the specific grouping and group-wise learning rate scaling strategies in SGG.

\subsection{Gradient Grouping via Online Clustering}
\label{sec:clustering}
It has been observed that parameters in LLMs exhibit non-independent optimization behaviors, inherently forming intra-correlated groups \citep{2024BOCB, zhang2025why_adam}. To build intuitions for this work, we first conduct an empirical analysis of gradient statistics with LLaMA pre-training on C4.

\vspace{-0.3em}
\paragraph{Pilot Studies.}
Figure~\ref{fig:sgg_clustering} shows that gradient statistics, whether measured by layers or parameters, exhibit distinct clustering patterns, which aligns with previous findings. However, a crucial aspect of these clusters is their \textit{internal diversity} -- they exhibit considerable parameter-wise variations within each group. Second, subtle yet significant deviation can be identified in these cluster distributions when examining different statistics, such as gradients in Figure~\ref{fig:cluster(a)} and learning rates in Figure~\ref{fig:cluster(b)}.

These findings lead to the following considerations: \textbf{(\romannumeral1)} Since the clustering patterns differ across optimization statistics (\textit{e.g.}, gradients vs. learning rates), methods relying on pre-defined fixed groups, such as Adam-mini~\citep{zhang2024adam_mini}, might not effectively capture these distinct behaviors, suggesting the need for dynamic grouping strategies.
\textbf{(\romannumeral2)} While grouping has proven effective, replacing parameter-wise learning rates simply with a single, aggregated one per group (either pre-defined or dynamically derived) might not adapt to the observed parameter-wise variation, thus discarding essential optimization signals for effective LLM training.

To this end, we propose to \textit{scale learning rates through dynamic grouping} rather than replace them in static groups, thereby imposing group constraint while maintaining parameter-wise adaptation.

\vspace{-0.4em}
\paragraph{Online Clustering.}
SGG begins with dynamic grouping as $\text{GradCluster}(m_l^t, K)$ in Algorithm~\ref{alg:sgg}, which partitions momentum vectors $m^t_l$ within each layer $l \in L$ into $K$ groups with related indices $\mathcal{C}_l^{t}$ according to their similarity. To achieve this, online clustering stands out as a straightforward solution, and the choice of specific clustering algorithms is then crucial for both effectiveness and efficiency. As such, we evaluate several potential strategies, including K-means, mini-batch K-means, Gaussian Mixture Models (GMM), and DBSCAN. Ablation studies in Figure~\ref{fig:ab_sgg_clustering} (perplexity versus training time) and Table~\ref{tab:ablation} (hyper-parameters) show that mini-batch K-means offer the most favorable trade-off between clustering quality and computational efficiency. Thus, we select this as the default clustering implementation of $\text{GradCluster}(m_l^t, K)$ in SGG.

\subsection{Cluster-specific Learning Rate Scaling}
\label{sec:scaling}
We introduce $\text{ScaleUpdate}(\mathcal{C}_l^{t}, g_l^t, \beta_3)$ to calculate the scaling factor $S_l^t [c]$ for each cluster $c \in K$ after grouping, which modulates learning rate $\alpha^t_l$. This involves two sub-tasks: \textbf{(\romannumeral1)} measuring the statistics for different levels of partitions (\textit{e.g.}, clusters, layers, and even the global one); \textbf{(\romannumeral2)} updating cluster-specific scales $S_l^t [c]$ based on the above statistics. This contrasts with the previous Adam-mini~\citep{zhang2024adam_mini}, which replaces per-parameter adaptive learning rates with their group-wise means directly.

\begin{table}[htb]
    \centering
    \vspace{-0.4em}
    \caption{\textbf{Group Statistics for SGG Scaling.} Param. refers to Adam-like baselines. Validation PPL$\downarrow$ is reported with LLaMA on C4. MDA yields the best result. 
    }
    \vspace{-0.70em}
    \setlength{\tabcolsep}{0.7mm}
\resizebox{1.0\linewidth}{!}{
    \begin{tabular}{l|gccc|c|cb}
    \toprule
Statistic & Var.   & Var.  & Sign(Var.) & Grad.      & Grad. & Grad.       \\
Method    & Param. & Mean  & Mean      & $L_2$-norm  & MAD   & \bf{MDA}    \\ \hline
130M      & 25.08  & 22.76 & 22.67      & 22.58      & 24.62 & \bf{22.18}  \\
1B        & 15.56  & 14.63 & 14.68      & 14.66      & 14.58 & \bf{14.30}  \\
    \bottomrule
    \end{tabular}
    }
    \label{tab:ab_sgg_statistic}
    \vspace{-0.5em}
\end{table}

To determine an effective measure for each cluster $c$ at layer $l \in L$, we examine several candidates in Table~\ref{tab:ab_sgg_statistic}.  These include: \textit{(1) Within-cluster metrics:} Mean in Adam-mini~\citep{zhang2024adam_mini}, Variance, Sign of Variance in SignSGD~\citep{icml2018signSGD}, $L_2$-norm in LARS~\citep{iclr2018lars}, and \textit{\textbf{(2)} Layer-aware metrics:} Median Absolute Deviation (MAD). More importantly, recent studies show that severe discrepancies exist between the training dynamics of shallow and deep LLM layers, resulting in sudden loss spikes~\citep{chowdhery2023palm, Molybog2023LossSpike}, exploding/vanishing gradients~\citep{wang2024deepnet, zhu2025DyT}, where the model's performance can deteriorate dramatically. This inspires us to incorporate a global perspective, beyond groups and layers, into SGG's scaling process to promote training homogeneity.

To achieve this, we adopt \textit{Median of Deviation to Average} (MDA). For each cluster $c\in K$ in layer $l$ at iteration $t$, its MDA, denoted $\mathcal{D}_{l,c}^{t}$, quantifies the median deviation of its constituent momentum $m_l^t \cdot \mathcal{C}_l^t[c]$ from the average of that layer, as:
\begin{equation}
\begin{aligned}
    \hspace{-1.25em}
    \mathcal{D}_{l,c}^{t} = \text{median} \left( |m_l^t \cdot \mathcal{C}_l^t[c] - \text{mean}(m_l^t)| \right),
\end{aligned}
\end{equation}
where \( \mathcal{C}_l^{t}[c] \) denotes the selection mask for parameters in cluster $c$ of layer $l$. To obtain a robust reference for homogenization, we compute a global MDA $\mathcal{D}^{t}$ which characterizes the typical parameter-wise deviation throughout the model. The scaling factor \( S_l[c] \) of cluster $c$ is then defined as the ratio of this global $\mathcal{D}^{t}$ to the cluster's specific $\mathcal{D}_{l,c}^{t}$ as:
\begin{equation}
    S_l^{t}[c] = \frac{\mathcal{D}^t}{\mathcal{D}_{l,c}^{t} + \epsilon},
\end{equation}
where $\epsilon=10^{-8}$ ensures numerical stability. Thus, clusters with a lower $\mathcal{D}_{l,c}^{t}$ (more stable dynamics relative to global $\mathcal{D}^t$) receive a proportionally larger factor $S_l^{t}[c]$. Conversely, clusters with higher, divergent MDAs are scaled down. Hence, this promotes learning homogeneity across layers and clusters, which mitigates discrepancies and suppresses divergent behavior that could lead to disruptive updates. 

\begin{table}[htb]
    \centering
    \vspace{-0.25em}
    \caption{\textbf{Gains vs Costs.} Relative \gbf{gains}$\uparrow$ in PPL and \rbf{cost}$\downarrow$ in training time and peak GPU memory increase for GPU, CPU, hybrid versions with LLaMA-1B on C4.
    }
    \vspace{-0.80em}
    \setlength{\tabcolsep}{0.7mm}
\resizebox{1.0\linewidth}{!}{
    \begin{tabular}{l|ccc}
    \toprule
Method                        & PPL                          & Training Time              & Memory             \\ \hline
\grow{Adam}                   & 15.56                        & 110h                       & 7.8G               \\
Adam+\bf{SGG} (GPU)           & \gbf{+6.5\%} \small{(-1.00)} & \rbf{+1.8\%} \small{(+2h)} & \rbf{+4.3G}        \\
Adam+\bf{SGG} (CPU)           & \gbf{+6.5\%} \small{(-1.00)} & \rbf{+8.2\%} \small{(+9h)} & \gray{\bf{+0.0G}} \\
\brow{Adam+\bf{SGG} (Hybrid)} & \gbf{+6.5\%} \small{(-1.00)} & \rbf{+4.1\%} \small{(+4h)} & \rbf{+2.1G}        \\
    \bottomrule
    \end{tabular}
    }
    \label{tab:ab_sgg_time}
    \vspace{-0.25em}
\end{table}

These factors are then clamped to $[0.1, 10]$ and are updated periodically per $T$ iterations using an EMA to smooth out short-term fluctuations:
\begin{equation}
    S_l^{t}[c] = \beta_3 \cdot S_l^{t-1}[c] + (1-\beta_3) \cdot \frac{\mathcal{D}^t}{\mathcal{D}_{c,l}^{t} + \epsilon},
\end{equation}
where $\beta_3$ is the EMA decay rate. Subsequently, per-parameter adaptive learning rates are multiplied by their corresponding group-wise scaling factors as $\alpha_l^t \cdot \mathcal{S}_l^{t}[\mathcal{C}_l^{t}]$ in Algorithm~\ref{alg:sgg}. Table~\ref{tab:ab_sgg_statistic} shows the effectiveness of using MDA for global homogenization while maintaining parameter-wise adaptation (\textit{e.g.}, 22.18 and 14.30 PPL for 130M and 1B models).

\begin{figure}[t!]
    \vspace{-0.50em}
    \centering
    \includegraphics[width=0.99\linewidth]{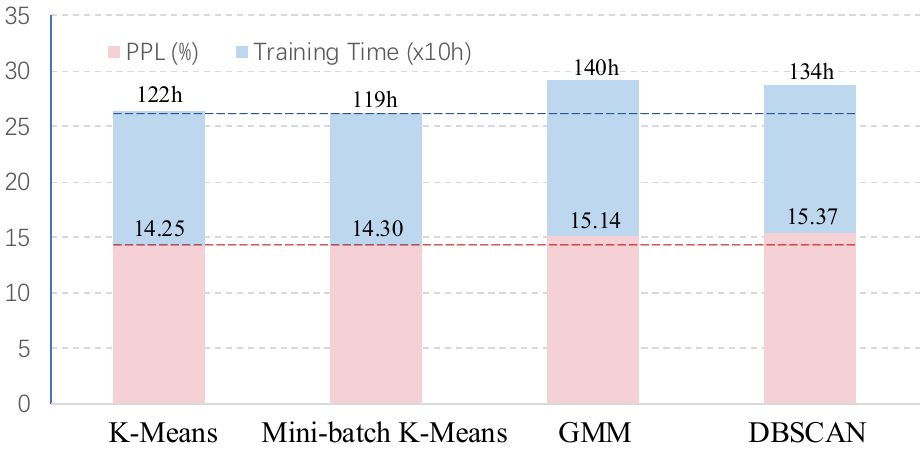}
    \vspace{-1.75em}
    \caption{\textbf{Grouping Methods PPL-efficiency trade-off} with LLaMA-1B on C4. Blue bars show validation Perplexity (PPL$\downarrow$), and pink bars show training time. Mini-batch K-means achieves the best trade-off.
    }
    \label{fig:ab_sgg_clustering}
    \vspace{-0.75em}
\end{figure}

As for implementation, we consider two key trade-offs between performance and efficiency.
\textbf{(\romannumeral1) Clustering Strategies and Frequency:}
We evaluate four common approaches: K-means~\citep{macqueen1967kmeans}, mini-batch K-means~\citep{www2010WebKMeans}, GMM~\citep{nips1994gmm}, and DBSCAN~\citep{kddm1996DBSCAN}. As depicted in Figure~\ref{fig:ab_sgg_clustering}, mini-batch K-means offers the best balance between accuracy and computational cost. We therefore adopt it as our default clustering strategy. We also empirically set the interval $T$ as 10\% of the total training iterations, as verified in Figure~\ref{fig:ablation_hyper}.
\textbf{(\romannumeral2) Storage and Computation:} As shown in Table~\ref{tab:ab_sgg_time}, we compare the performance, training time, and GPU memory of putting them on the GPU or CPU.
While keeping $\{\mathcal{C}_l\}_{l=1}^{L}$ and $\{\mathcal{S}_l\}_{l=1}^{L}$ in CPU would not slow the training, the peak GPU memory for online clustering is significant. 
Consequently, as stated in Table~\ref{tab:intro}, we utilize the CPU implementation that stores additional optimization states on the CPU, which does not require extra GPU memory and increases the overall training time negligibly.
\section{Experiments}
\label{sec:exp}

\begin{figure*}[t]
    \vspace{-0.5em}
    \centering
    \subfigtopskip=-0.5pt
    \subfigbottomskip=-0.5pt
    \subfigcapskip=-4.0pt
    \subfigure[LLaMA-130M Pre-training]{\hspace{-0.25em}
    \label{fig:c4(a)}\includegraphics[width=0.32\linewidth,trim= 0 0 0 0,clip]{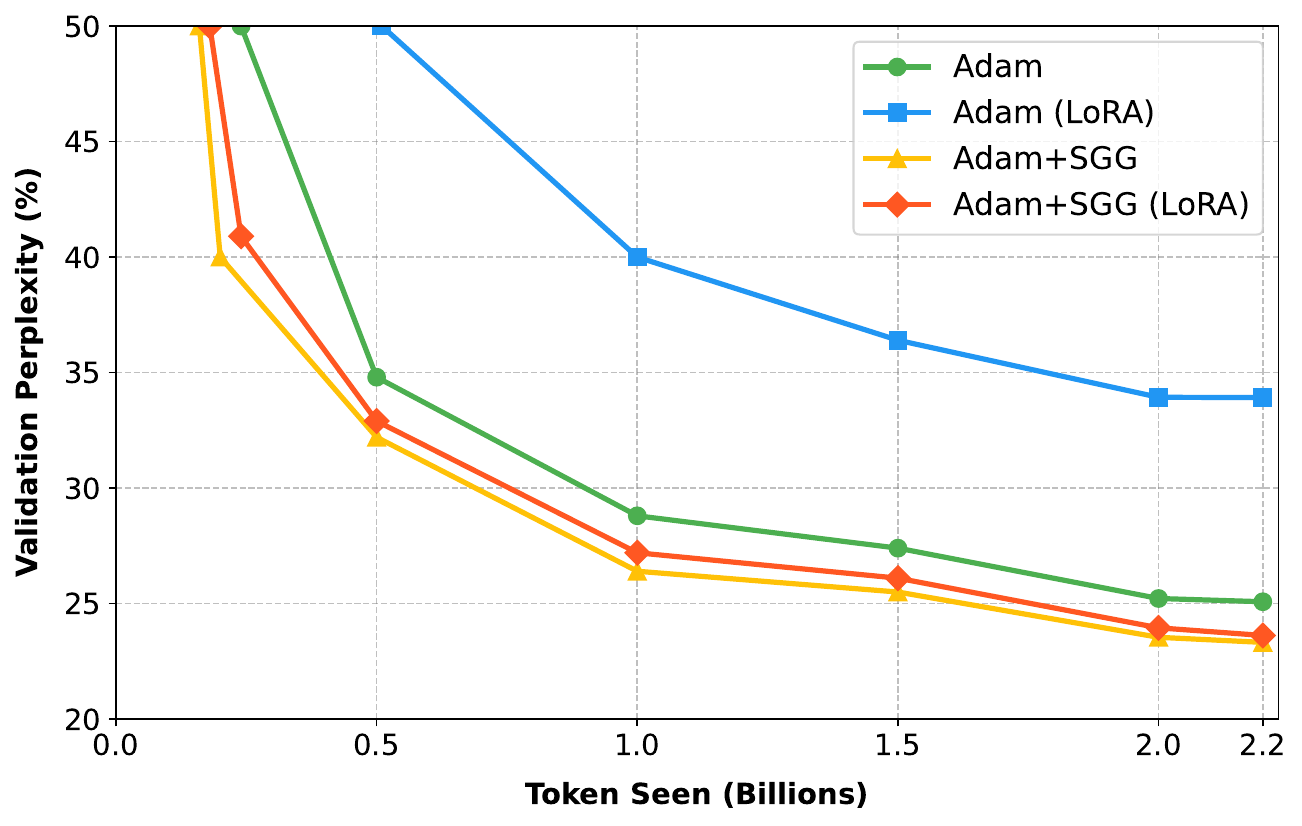}}
    \subfigure[LLaMA-1B Pre-training]{\hspace{-0.25em}
    \label{fig:c4(b)}\includegraphics[width=0.32\linewidth,trim= 0 0 0 0,clip]{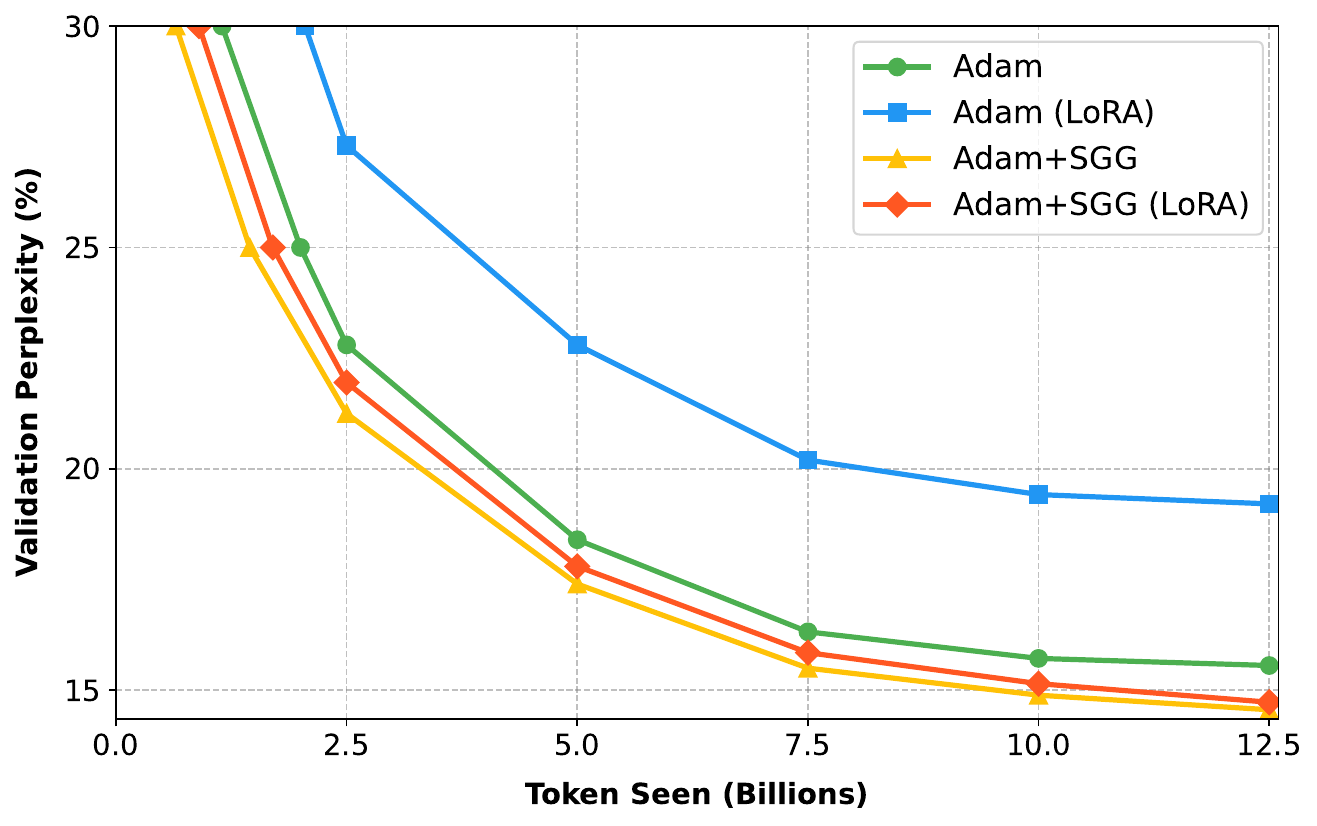}}
    \subfigure[Parameter Scaling-up]{\hspace{-0.25em}
    \label{fig:c4(c)}\includegraphics[width=0.33\linewidth,trim= 0 0 0 0,clip]{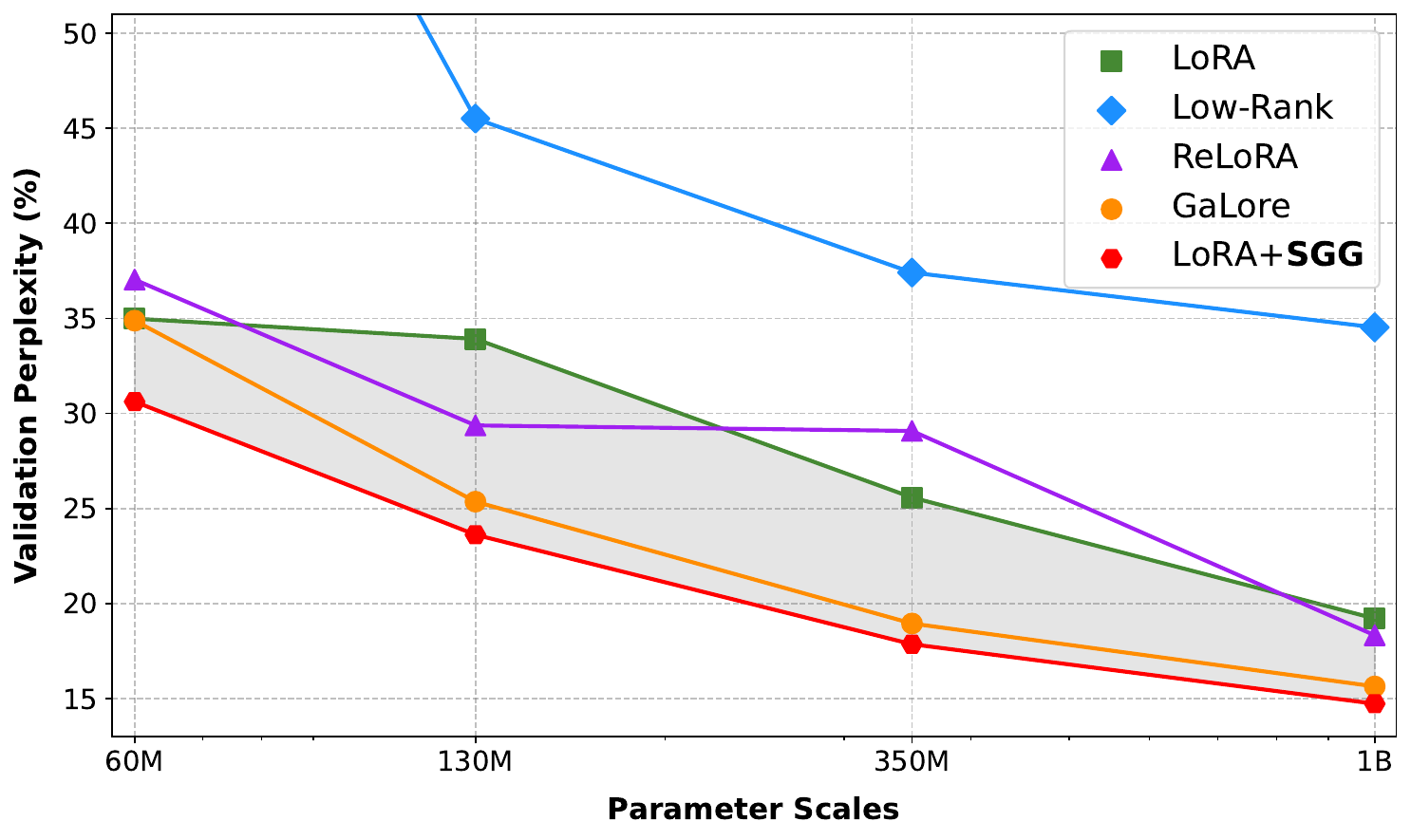}}
    \vspace{-0.60em}
    \caption{\textbf{Convergence and Scaling-up on C4 Pre-training}. Validation Perplexity (PPL\%$\downarrow$: lower is better) vs Training Tokens/Parameters.
    \textbf{(a)} LLaMA-130M and \textbf{(b)} LLaMA-1B training curves demonstrate faster convergence and lower PPL of SGG compared to baselines in both low-rank (\textcolor{fig3blue}{\textbf{Adam}} vs \textcolor{fig3red}{\textbf{Adam+SGG}}) and full-rank (\textcolor{fig3green}{\textbf{Adam}} vs \textcolor{fig3yellow}{\textbf{Adam+SGG}}) settings. \textbf{(c)} \textcolor{fig3red}{\textbf{LoRA+SGG}} consistently outperforms other low-rank methods as model size increases.
    }
    \label{fig:c4_scaling}
    \vspace{-0.5em}
\end{figure*}

\begin{table}[t]
    \centering
    \vspace{-0.25em}
    \caption{\textbf{C4 Pre-training with diverse LLaMA sizes} (from 60M to 1B). Comparison of full-rank, memory-efficient, and low-rank optimizers. Validation Perplexity (PPL\%$\downarrow$: lower is better) is reported. \textbf{Bold} and \gbf{green} types denote the best results and gains$\downarrow$ of SGG (\sethlcolor{lightblue}\hl{blue background}) over related baselines (\sethlcolor{gray90}\hl{gray background}). Note that $\dagger$ denotes the results borrowed from GaLore, while the others were reimplemented in this work.
    }
    \vspace{-0.70em}
    \setlength{\tabcolsep}{0.8mm}
\resizebox{1.0\linewidth}{!}{
    \begin{tabular}{lc|cccc}
    \toprule
\textbf{Method}              & \textbf{Venue}       & \textbf{60M} & \textbf{130M} & \textbf{350M} & \textbf{1B} \\ \hline
\multicolumn{6}{l}{ \gray{\textit{Pre-training with Full-Rank Optimizers}}} \\
\grow Adam$^\dag$            & ICLR'15~             & ~34.06       & 25.08         & 18.80         & 15.56       \\
NAdam                        & ICLR'18~             & ~35.86       & 28.88         & 19.24         & 15.78       \\
RAdam                        & ICLR'20~             & ~30.43       & 25.17         & 19.13         & 15.65       \\
LAMB                         & ICLR'20~             & ~33.04       & 24.37         & 18.26         & 15.84       \\
Adan                         & TPAMI'23~            & ~32.01       & 23.14         & 17.32         & 14.70       \\
\brow Adam+\bf{SGG}          & \bf{Ours}            & \bf{30.31}   & \bf{22.18}   & \bf{17.28}    & \bf{14.30}   \\
\multicolumn{2}{l|}{\gray{$\Delta$ \textit{Gains}}} & \gbf{-3.75}  & \gbf{-2.90}   & \gbf{-1.52}   & \gbf{-1.26} \\
\hline
\multicolumn{6}{l}{ \gray{\textit{Pre-training with Memory-efficient Optimizers}}} \\
Adam-mini$^\dag$             & ICLR'25~             & ~34.10       & 24.85         & 19.05         & 16.07       \\
Adafactor$^\dag$             & ICML'18~             & 32.57        & 23.98         & 17.74         & 15.19       \\
Low-Rank$^\dag$              & arXiv'22~            & 78.18        & 45.51         & 37.41         & 34.53       \\
\grow CAME                   & ACL'23~              & 31.37        & 23.38         & 17.45         & 14.68       \\
\brow CAME+\bf{SGG}          & \bf{Ours}            & \bf{30.15}   & 22.91         & 17.09         & 14.35       \\
\multicolumn{2}{l|}{\gray{$\Delta$ \textit{Gains}}} & \gbf{-1.22}  & \gbf{-0.46}   & \gbf{-0.36}   & \gbf{-0.33} \\
\grow APOLLO$^\dag$          & MLSys'25             & 31.55        & 22.94         & 16.85         & 14.20       \\
\brow APOLLO+\bf{SGG}        & \bf{Ours}            & 30.18        & \bf{22.52}    & \bf{16.54}    & \bf{13.95}  \\
\multicolumn{2}{l|}{\gray{$\Delta$ \textit{Gains}}} & \gbf{-1.37}  & \gbf{-0.42}   & \gbf{-0.31}   & \gbf{-0.25} \\ \hline
\multicolumn{6}{l}{ \gray{\textit{Low-Rank Pre-training}}} \\
\grow LoRA$^\dag$            & ICLR'22              & 34.99        & 33.92         & 25.58         & 19.21       \\
ReLoRA$^\dag$                & ICLR'23              & 37.04        & 29.37         & 29.08         & 18.33       \\
GaLore$^\dag$                & ICML'24              & 34.88        & 25.36         & 18.95         & 15.64       \\
\brow LoRA+\bf{SGG}          & \bf{Ours}            & \bf{30.62}   & \bf{23.62}    & \bf{17.86}    & \bf{14.73}  \\
\multicolumn{2}{l|}{\gray{$\Delta$ \textit{Gains}}} & \gbf{-4.37}  & \gbf{-10.30}  & \gbf{-7.72}   & \gbf{-4.48} \\ \hline
\multicolumn{2}{l|}{Training Tokens}                & 1.1B         & 2.2B          & 6.4B          & 13.1B       \\
    \bottomrule
    \end{tabular}
    }
    \label{tab:comp_c4_pt}
    \vspace{-0.75em}
\end{table}

\subsection{Experimental Setup}
\label{sec:exp_setup}
\paragraph{Datasets and Tasks.}
To evaluate the effectiveness and versatility of SGG, we conducted experiments on 20 public datasets, including large-scale natural language datasets, Visual Question Answering (VQA), and multimodal LLM (MLLM) evaluation benchmarks.
\textbf{(1) Pre-training on C4:} We used the \texttt{en} subset of the C4 dataset, a large cleaned web corpus from Common Crawl filtered for safety~\cite{kopf2023openassistant}, to assess SGG in LLM pre-training.
\textbf{(2) SFT on GLUE:} We fine-tuned RoBERTa-base models on GLUE benchmark. GLUE comprises a collection of NLP tasks, such as sentiment analysis, question answering, and textual entailment~\cite{wang2018glue}, providing a standard measurement of generalization in the understanding capabilities of common languages.
\textbf{(3) PEFT on Commonsense Reasoning:} Leveraging the LLM-Adapters framework~\cite{hu2023llmadapters}, we evaluated SGG's compatibility and performance with PEFT methods on LLaMA architecture across 8 Commonsense Reasoning (CS) datasets: BoolQ~\cite{clark2019boolq}, PIQA~\cite{bisk2020piqa}, SIQA~\cite{sap2019socialiqa}, HellaSwag~\cite{zellers2019hellaswag}, WinoGrande~\cite{sakaguchi2021winogrande}, ARC (ARC-Easy and ARC-Challenge)~\cite{clark2018arc}, and OBQA~\cite{mihaylov2018obqa}.
\textbf{(4) Direct Preference Optimization (DPO):} To evaluate SGG in human preference alignment tasks, we implemented DPO using the TRL library. The Qwen2.5 0.5B model was trained on the \texttt{ultrafeedback\_binarized} dataset, which includes binary preference labels~\cite{vonwerra2022trl}.
\textbf{(5) MLLM Validation:} (i) VQA benchmarks such as GQA~\cite{hudson2019gqa}, TextVQA~\cite{singh2019textvqa}, SciVQA$^I$ (evaluation on the imageset of ScienceVQA)~\cite{lu2022scivqa}, VQAv2~\cite{goyal2017vqav2}, and Vizwiz~\cite{gurari2018vizwiz}. (ii) MLLM evaluation benchmarks including POPE~\cite{li2023pope}, MMBench~\cite{liu2025mmbench}, MMBench-Chinese (MMBench$^{\text{CN}}$)~\cite{liu2025mmbench}, SEED$^I$~\cite{li2023seed}, and MME (Perception)~\cite{yin2023mme}.

\vspace{-0.25em}
\paragraph{Implementation Details}
We implemented SGG in PyTorch, ensuring compatibility with standard optimizers through minimal code integration. Its key hyper-parameters were empirically tuned for optimal performance-efficiency trade-off as the default setups: cluster number $K=3$, interval $T=500$ (nearly 1$\sim$5$\%$ of total iterations), and decay coefficient \(\beta_3 = 0.99\). To minimize GPU memory demands, cluster indices $\mathcal{C}$ and scaling factors $\mathcal{S}$ can be optionally stored on the CPU. Table~\ref{tab:ab_sgg_time} confirms SGG's negligible training time increase and preserved GPU memory footprint.
Reproduced results are marked with \sethlcolor{gray90}\hl{gray} and \sethlcolor{lightblue}\hl{blue} backgrounds, while the others are cited from their original papers. All experiments are conducted using NVIDIA A100-80G GPUs with three independent runs.

\begin{table*}[t]
    \centering
    \vspace{-0.5em}
    \caption{\textbf{GLUE Benchmark Results with RoBERTa-base.} Top-1 accuracy (\%$\uparrow$: higher is better) is reported. Comparison across both full-rank and low-rank (LoRA $r=4$, $r=8$) settings. \textbf{Bold} and \gbf{green} types denote the best results and  performance gains$\uparrow$ of SGG (\sethlcolor{lightblue}\hl{blue background}) compared to related baselines (\sethlcolor{gray90}\hl{gray background}).
    }
    \vspace{-0.70em}
    \setlength\tabcolsep{0.25em}
    \resizebox{1.0\linewidth}{!}{
        \begin{tabular}{lc|rlrlrlrlrlrlrlrl|rl}
        \toprule
\bf{Optimizer}                 & \bf{Rank}         & \multicolumn{2}{c}{\bf{CoLA}} & \multicolumn{2}{c}{\bf{STS-B}} & \multicolumn{2}{c}{\bf{MRPC}} & \multicolumn{2}{c}{\bf{RTE}} & \multicolumn{2}{c}{\bf{SST2}} & \multicolumn{2}{c}{\bf{MNLI}} & \multicolumn{2}{c}{\bf{QNLI}} & \multicolumn{2}{c|}{\bf{QQP}} & \multicolumn{2}{c}{\bf{Average}} \\ \hline
\multicolumn{2}{l}{\gray{\textit{Full-Rank SFT}}} &            &                  &            &                   &            &                  &           &                  &            &                  &            &                  &            &                  &            &                  &              &                   \\
SGD                            & Full              &~62.12      &                  & 90.73      &                   & 87.74      &                  & 79.06     &                  & 94.26      &                  & 87.53      &                  & 92.29      &                  & 92.22      &                  & 85.74        &                   \\
\grow AdamW                    & Full              &~62.24      &                  & 90.92      &                   & 91.30      &                  & 79.42     &                  & 94.57      &                  & 87.18      &                  & 92.33      &                  & 92.28      &                  & 86.24        &                   \\
\grow LAMB                     & Full              &~62.09      &                  & 90.59      &                   & 88.72      &                  & 75.45     &                  & 94.72      &                  & 87.71      &                  & 92.42      &                  & 91.46      &                  & 85.40        &                   \\
CAME                           & Full            &~62.16     &                 & 90.43     &                  & 89.02     &                 & 75.94    &                 & 94.61     &                 & 87.13     &                 & 92.31     &                 & 91.54     &                 & 85.39        &                 \\
APOLLO                         & Full            &~62.45     &                 & 90.70     &                  & 90.36     &                 & 77.53    &                 & 94.58     &                 & 87.57     &                 & 92.40     &                 & 92.12     &                 & 85.96        &                 \\
\brow{AdamW}+\bf{SGG}         & Full             &~63.36        & \gbf{+1.12}~   & \bf{91.22}    & \gbf{+0.30}~   & \bf{92.65}   & \gbf{+1.35}~   & \bf{80.87}   & \gbf{+1.45}~  & \bf{95.58}   & \gbf{+1.01}~   & \bf{88.32}   & \gbf{+1.14}~   & 92.88        & \gbf{+0.55}~   & \bf{93.32}   & \gbf{+1.04}~   & \bf{87.28}     & \gbf{+1.00}     \\
\brow{LAMB}+\bf{SGG}          & Full             &~62.47        & \gbf{+0.38}~   & 90.90         & \gbf{+0.31}~   & 89.46        & \gbf{+0.74}~   & 76.53        & \gbf{+1.08}~  & 94.95        & \gbf{+0.23}~   & 87.81        & \gbf{+0.10}~   & \bf{92.89}   & \gbf{+0.47}~   & 91.78        & \gbf{+0.32}~   & 85.85          & \gbf{+0.45}     \\
\hline
\multicolumn{2}{l}{\gray{\textit{Low-Rank SFT (rank 4)}}}  &            &                  &            &                   &            &                  &           &                  &            &                  &            &                  &            &                  &            &                  &              &                   \\
SGD (LoRA)                     & 4                 & 60.32      &                  & 90.31      &                   & 87.75      &                  & 79.06     &                  & 94.27      &                  & 87.39      &                  & 92.16      &                  & 91.89      &                  & 85.39        &                   \\
\grow AdamW (LoRA)             & 4                 & 61.38      &                  & 90.57      &                   & 91.07      &                  & 78.70     &                  & 92.89      &                  & 86.82      &                  & 92.18      &                  & 91.29      &                  & 85.61        &                   \\
\grow LAMB (LoRA)              & 4                 & 61.51      &                  & 90.33      &                   & 89.46      &                  & 74.73     &                  & 94.27      &                  & 87.51      &                  & 92.48      &                  & 91.57      &                  & 85.23        &                   \\
DoRA                           & 4                 & 60.38      &                  & 90.50      &                   & 88.24      &                  & 74.73     &                  & 93.69      &                  & $-$        &                  & 92.59      &                  & $-$        &                  & $-$          &                   \\
GaLore (LoRA)                  & 4                 & 60.35      &                  & 90.73      &                   & 92.25      &                  & 79.42     &                  & 94.04      &                  & 87.00      &                  & 92.24      &                  & 91.06      &                  & 85.89        &                   \\
\brow{AdamW}+\bf{SGG}         & 4                & 62.36        & \gbf{+0.98}~   & \bf{91.10}    & \gbf{+0.53}~   & \bf{92.12}        & \gbf{+1.05}~   & \bf{80.51}   & \gbf{+1.81}~  & \bf{95.06}   & \gbf{+2.17}~   & \bf{88.18}   & \gbf{+1.36}~   & 92.62        & \gbf{+0.44}~   & \bf{93.06}   & \gbf{+1.77}~   & \bf{86.88}     & \gbf{+1.27}     \\
\brow{LAMB}+\bf{SGG}          & 4                & \bf{62.47}   & \gbf{+0.96}~   & 90.90         & \gbf{+0.57}~   & 89.46        & \gbf{+0.30}~   & 75.53        & \gbf{+0.80}~  & 94.95        & \gbf{+0.34}~   & 87.73        & \gbf{+0.12}~   & \bf{92.92}   & \gbf{+0.41}~   & 91.78        & \gbf{+0.36}~   & 85.72          & \gbf{+0.49}     \\
\hline
\multicolumn{2}{l}{\gray{\textit{Low-Rank SFT (rank 8)}}}  &            &                  &            &                   &            &                  &           &                  &            &                  &            &                  &            &                  &            &                  &              &                   \\
SGD (LoRA)                     & 8                 & 60.57      &                  & 90.29      &                   & 88.48      &                  & 79.42     &                  & 94.32      &                  & 87.44      &                  & 92.23      &                  & 92.10      &                  & 85.61        &                   \\
\grow AdamW (LoRA)             & 8                 & 61.83      &                  & 90.80      &                   & 91.90      &                  & 79.06     &                  & 93.46      &                  & 86.94      &                  & 92.25      &                  & 91.22      &                  & 85.93        &                   \\
\grow LAMB (LoRA)              & 8                 & 61.89      &                  & 90.78      &                   & 89.21      &                  & 79.42     &                  & 94.61      &                  & 87.61      &                  & 92.51      &                  & 91.42      &                  & 85.35        &                   \\
DoRA                           & 8                 & 58.36      &                  & 90.63      &                   & 88.97      &                  & 75.09     &                  & 93.81      &                  & $-$        &                  & 92.68      &                  & $-$        &                  & $-$          &                   \\
GaLore (LoRA)                  & 8                 & 60.06      &                  & 90.82      &                   & 92.01      &                  & 79.78     &                  & 94.38      &                  & 87.17      &                  & 92.20      &                  & 91.11      &                  & 85.94        &                   \\
\brow{AdamW}+\bf{SGG}         & 8                & 62.36        & \gbf{+0.53}~   & \bf{91.10}    & \gbf{+0.30}~   & \bf{92.12}   & \gbf{+0.22}~   & 80.51        & \gbf{+1.45}~  & \bf{95.06}   & \gbf{+1.60}~   & \bf{88.17}   & \gbf{+1.23}~   & 92.65        & \gbf{+0.40}~   & \bf{92.85}   & \gbf{+1.63}~   & \bf{86.85}     & \gbf{+0.92}     \\
\brow{LAMB}+\bf{SGG}          & 8                & \bf{62.47}   & \gbf{+0.58}~   & 90.90         & \gbf{+0.12}~   & 89.46        & \gbf{+0.25}~   & 76.53        & \gbf{+1.80}~  & 94.95        & \gbf{+0.34}~   & 87.85        & \gbf{+0.24}~   & \bf{92.87}   & \gbf{+0.36}~   & 91.78        & \gbf{+0.36}~   & 85.85          & \gbf{+0.50}     \\
        \bottomrule
    \end{tabular}
    \label{table:comp_GLUE_full}
    \vspace{-1.0em}
}
\end{table*}

\subsection{Comparison Results with LLMs}
\label{sec:comp_llm}
Across PT, SFT, PEFT, and DPO, SGG consistently improves performance with efficiency, highlighting its value as a versatile optimizer wrapper for LLMs.

\paragraph{Pre-training on C4.}
Following GaLore~\citep{zhao2024galore}, we employ LLaMA-based architectures (60M to 1B) for both full-rank and low-rank pre-training. We keep consistent hyper-parameters, tuning learning rates within a fixed budget, and use BF16 precision for efficiency. Table~\ref{tab:comp_c4_pt} shows that applying SGG consistently reduces validation perplexity (\gbf{-3.75\%} and \gbf{-1.26\%} for AdamW in 60M and 1B; \gbf{-10.30\%} and \gbf{-4.48\%} for LoRA in 130M and 1B) and it accelerates convergence (Figure~\ref{fig:c4_scaling}) compared to baselines. Notably, SGG for the first time \textit{enables low-rank pre-training (LoRA+SGG) to achieve performance comparable to full-rank training across model sizes} (\textit{e.g.}, \textbf{14.73} vs \textbf{14.30} in 1B; \textbf{30.62} vs \textbf{30.31} in 60M), a huge step forward as previous low-rank optimizers typically lagged behind in performance. View Appendix~\ref{app:implement} for details.

\paragraph{SFT on GLUE.}
We fine-tuned the pre-trained RoBERTa-base on various GLUE tasks. Table~\ref{table:comp_GLUE_full} shows that applying SGG yields consistent gains over baselines in both full-rank and low-rank (ranks 4 and 8) SFT scenarios. Notably, AdamW+SGG yields substantial average gains (\gbf{+1.00\%} full-rank, \gbf{+1.27\%} rank 4), with significant task-specific improvements (\textit{e.g.}, MRPC full-rank \gbf{+1.35\%}, MNLI rank 4 \gbf{+1.36\%}), demonstrating SGG's versatility and robustness across different SFT scenarios.

\paragraph{PEFT on Commonsense Reasoning.}
Following LLM-Adapters, we assess SGG in CS tasks with top-1 accuracy and GPU memory, where LLaMA-7B is fine-tuned by AdamW+LoRA ($r=32$) on a unified training dataset, followed by evaluation on each specific subset. As shown in Table~\ref{tab:comp_cs_avg}, SGG improves LoRA by an average of \gbf{+2.9\%} across all tasks, with up to \gbf{+4.2\%} gains on specific tasks like OBQA. It matches or surpasses PEFT baselines, such as Prefix~\citep{Li2021Prefix}, Series\citep{Houlsby2019Series}, and Parallel~\citep{He2021Parallel}, and more recent DoRA, GaLore, and Fira~\citep{chen2024fira}. View Table~\ref{tab:comp_cs_full} and Appendix~\ref{app:implement} for details.

\begin{table}[t!]
    \centering
    \vspace{-0.5em}
    \caption{\textbf{LLaMA-7B PEFT Results} on Commonsense Reasoning. Comparison of LoRA+SGG (\sethlcolor{lightblue}\hl{blue background}) against baselines. Top-1 accuracy (\%$\uparrow$: higher is better) of selected tasks and all tasks on average (Avg.) are reported. \textbf{Bold} and \gbf{green} types denote the best results and gains$\uparrow$ compared to LoRA (\sethlcolor{gray90}\hl{gray background}).
    }
    \vspace{-0.70em}
    \setlength{\tabcolsep}{0.8mm}
\resizebox{1.0\linewidth}{!}{
    \begin{tabular}{l|cccccc|c}
    \toprule
\textbf{Method}               & \textbf{BoolQ} & \textbf{PIQA} & \textbf{SIQA} & \textbf{WG}   & \textbf{Arc-E} & \textbf{OBQA} & \textbf{Avg.}  \\ \hline
Parallel                      & 67.9           & 76.4          & 78.8          & 78.9          & 73.7           & 75.2          & 72.2           \\
\grow{LoRA}                   & 68.9           & 80.7          & 77.4          & 78.8          & 77.8           & 74.8          & 74.7           \\
DoRA                          & 69.7           & 83.4          & 78.6          & 81.0          & 81.9           & 79.2          & \textbf{78.4}  \\
GaLore                        & 69.5           & 82.0          & 75.1          & 18.0          & 80.7           & 78.0          & 62.7           \\
Fira                          & 69.4           & 82.6          & 78.0          & \textbf{81.2} & \textbf{82.2}  & \textbf{80.8} & 76.9           \\ \hline
\brow{LoRA}+\bf{SGG}          & \textbf{70.3}  & \textbf{83.6} & \textbf{78.8} & 80.9          & 81.5           & 79.0          & 77.6           \\
\gray{$\Delta$ \textit{Gains}}& \gbf{+1.4}     & \gbf{+2.9}    & \gbf{+1.4}    & \gbf{+2.1}    & \gbf{+3.7}     & \gbf{+4.2}    & \gbf{+2.9}     \\ \hline
\brow{DoRA}+\bf{SGG}          & \textbf{71.4}  & \textbf{84.8} & \textbf{79.5} & 82.8          & 83.8           & 81.2          & 79.6           \\
\gray{$\Delta$ \textit{Gains}}& \gbf{+1.7}     & \gbf{+1.4}    & \gbf{+0.9}    & \gbf{+1.8}    & \gbf{+1.9}     & \gbf{+2.0}    & \gbf{+1.2}     \\
    \bottomrule
    \end{tabular}
    }
    \label{tab:comp_cs_avg}
    \vspace{-1.0em}
\end{table}

\begin{table}[t]
    \centering
    \vspace{-0.15em}
    \caption{\textbf{Qwen2.5-0.5B DPO Results} with full-rank and LoRA setups. Top-1 accuracy(\%)$\uparrow$ is reported. \textbf{Bold} and \gbf{green} types denote best results and relative gains.
    }
    \vspace{-0.60em}
    \setlength{\tabcolsep}{1.3mm}
\resizebox{0.82\linewidth}{!}{
    \begin{tabular}{l|rl|rl}
    \toprule
\textbf{Optimizer}      & \multicolumn{2}{c|}{\textbf{Full-Rank}}   & \multicolumn{2}{c}{\textbf{LoRA}}         \\ \hline
\gcell{SGD}             & ~\gcell{70.10}      & \gcell{}            & ~\gcell{69.73}      & \gcell{}            \\
\gcell{AdamW}           & ~\gcell{71.39}      & \gcell{}            & ~\gcell{70.22}      & \gcell{}            \\
\gcell{LAMB}            & ~\gcell{70.82}      & \gcell{}            & ~\gcell{70.39}      & \gcell{}            \\
\bcell{SGD+\bf{SGG}}~   & ~\bcell{70.82}      & \bcell{\gbf{+0.72}} & ~\bcell{70.76}      & \bcell{\gbf{+1.03}} \\
\bcell{AdamW+\bf{SGG}}~ & ~\bcell{\bf{71.85}} & \bcell{\gbf{+0.47}} & ~\bcell{\bf{72.02}} & \bcell{\gbf{+1.80}} \\
\bcell{LAMB+\bf{SGG}}~  & ~\bcell{71.32}      & \bcell{\gbf{+0.50}} & ~\bcell{71.28}      & \bcell{\gbf{+0.89}} \\
    \bottomrule
    \end{tabular}
    }
    \label{tab:comp_dpo}
    \vspace{-1.0em}
\end{table}
\begin{table}[t]
    \centering
    \vspace{-0.5em}
    \caption{\textbf{MLLM performance comparison} on diverse benchmarks with LLaVA variants and different optimizers. Top-1 accuracy (\%)$\uparrow$ for selected tasks and all-task averaged (Avg.) results are reported. MMB and MMB$^{\text{CN}}$ denote MMbench and MMbench (Chinese). \textbf{Bold} and \gbf{green} types denote the best results and gains$\downarrow$ of SGG (\sethlcolor{lightblue}\hl{blue background}) over related baselines (\sethlcolor{gray90}\hl{gray background}). Please view Table~\ref{table:comp_mllm_full} for the full results.
    }
    \vspace{-0.70em}
    \renewcommand{\arraystretch}{1.2}
    \setlength\tabcolsep{0.1em}
    \resizebox{1.0\linewidth}{!}{
        \begin{tabular}{l|cccc|cccc}
        \toprule
        \multirow{2}{*}{\bf{Optimizer}} & \multicolumn{4}{c}{\bf{Image Question Answering}} & \multicolumn{3}{c}{\bf{Benchmarks}} & \multirow{2}{*}{\bf{Avg.}} \\
        \cline{2-8}
                            & GQA  & VizWiz & SciVQA$^I$ & VQA$^T$  & MMB  & MMB$^{\text{CN}}$  & POPE &       \\ \hline
        BLIP-2              & 41.0 & 19.6   & 61.0       & 42.5     & $-$  & $-$                & 85.3 & $-$   \\
        InstructBLIP        & 49.2 & 34.5   & 60.5       & 50.1     & 36.0 & 23.7               & 79.8 & 47.7   \\
        Qwen-VL             & 59.3 & 35.2   & 67.1       & 63.8     & 38.2 & 7.4                & $-$  & $-$   \\
        TinyLLaVA           & 62.0 & $-$    & 69.1       & 59.1     & 66.9 & $-$                & 86.4 & $-$   \\
        MoE-LLaVA           & 62.6 & $-$    & \bf{70.3}  & 57.0     & 68.0 & $-$                & 85.7 & $-$   \\
        LLaVA-Phi           & $-$  & $-$    & 68.4       & 48.6     & 59.8 & $-$                & 85.0 & $-$   \\ 
        LLaVA-NeXT          & \bf{64.2} & \bf{57.6} & 70.1 & \bf{64.9} & \bf{67.4} & 60.6       & 86.5 & \bf{67.3}   \\
        LLaVA-MOD           & 58.7 & 39.2   & 68.0       & 58.5     & 66.3 & 61.9          & \bf{87.0} & 62.8   \\
        LLaVA-KD-2B         & 62.3 & 44.7   & 64.7       & 53.4     & 64.0 & \bf{63.7}          & 86.3 & 62.7   \\ \hline
        \multicolumn{9}{l}{ \gray{\textit{LLaVA-v1.5 Full-Rank SFT}}}  \\
\grow   AdamW               & 62.0     & 50.0      & 66.8      & \bf{58.2}      & 64.3      & 58.3           & 85.9      & 63.6     \\
\grow   Adafactor           & 62.7     & 48.2      & 70.7      & 57.1      & 66.1      & 60.4           & 86.0      & 64.5     \\
\grow   LAMB                & 43.8     & 53.3 & 61.5      & 43.4      & 43.2      & 41.8           & 81.2      & 52.6     \\
\brow   AdamW+\bf{SGG}      & 62.4     & 50.2      & 69.8      & 57.4 & 65.9      & 60.1           & \bf{86.3} & 64.6     \\ 
        \multicolumn{1}{l|}{ \gray{$\Delta$ \textit{Gains}}} & \gbf{+0.4} & \gbf{+0.2}  & \gbf{+3.0}  & \gbf{-0.8}  & \gbf{+1.6} & \gbf{+1.8} & \gbf{+0.4} & \gbf{+1.0} \\
\brow   Adafactor+\bf{SGG}  &\bf{62.8} & 50.6      & \bf{71.6} & 57.3      & \bf{66.3} & \bf{60.8}      & 86.0      & \bf{65.1} \\
        \multicolumn{1}{l|}{ \gray{$\Delta$ \textit{Gains}}} & \gbf{+0.1} & \gbf{+2.4}  & \gbf{+0.9}  & \gbf{+0.2}  & \gbf{+0.2} & \gbf{+0.4} & \gbf{+0.0} & \gbf{+0.6} \\
\brow   LAMB+\bf{SGG}       & 44.0 & \bf{53.3}      & 61.8 & 43.5      & 43.3 & 41.9      & 81.3      & 52.7 \\
        \multicolumn{1}{l|}{ \gray{$\Delta$ \textit{Gains}}} & \gbf{+0.2} & \gbf{+0.0}  & \gbf{+0.3}  & \gbf{+0.1}  & \gbf{+0.1} & \gbf{+0.1} & \gbf{+0.1} & \gbf{+0.1} \\ \hline
        \multicolumn{9}{l}{ \gray{\textit{LLaVA-v1.5 Low-Rank SFT (AdamW)}}}  \\
\grow   LoRA                & 63.0      & 47.8      & 68.4      & 58.2      & 66.1      & 58.9       & 86.4      & 64.1      \\
\brow   LoRA+\bf{SGG}       & \bf{63.4} & \bf{51.0} & \bf{70.1} & \bf{58.6} & \bf{66.7} & \bf{59.4}  & \bf{86.6} & \bf{65.1} \\
        \multicolumn{1}{l|}{ \gray{$\Delta$ \textit{Gains}}} & \gbf{+0.4} & \gbf{+2.2}  & \gbf{+1.5}  & \gbf{+0.4}  & \gbf{+0.6} & \gbf{+0.5} & \gbf{+0.2} & \gbf{+1.0} \\ \hline
        \multicolumn{9}{l}{ \gray{\textit{LLaVA-v1.5 8-bit Low-Rank SFT (AdamW)}}}  \\
\grow   Q-LoRA              & 54.3      & 50.7      & 66.4     & 52.5      & 56.0      & 49.8       & 82.9      & 58.9      \\
\brow   Q-LoRA+\bf{SGG}     & \bf{55.1} & \bf{51.3} & \bf{66.7} & \bf{53.0} & \bf{56.1} & \bf{51.0}  & \bf{83.4} & \bf{59.5} \\
        \multicolumn{1}{l|}{\gray{$\Delta$ \textit{Gains}}} & \gbf{+0.8} & \gbf{+0.6}  & \gbf{+0.3}  & \gbf{+0.5}  & \gbf{+0.1} & \gbf{+0.2} & \gbf{+0.5} & \gbf{+0.6} \\
%
    \bottomrule
    \end{tabular}
    \label{table:comp_mllm_main}
    \vspace{-0.75em}
}
\end{table}

\paragraph{DPO.}
We verify SGG's effectiveness in aligning LLMs with human preferences using DPO, adhering to standard TRL library settings. SGG again demonstrates clear advantages. As shown in Table~\ref{tab:comp_dpo}, AdamW+SGG achieves the highest accuracy (\gbf{72.02\%}) under LoRA training, improving significantly (\gbf{+1.80\%}) over AdamW and \textit{even surpassing its full rank counterpart} (\textbf{72.02\%} vs \textbf{71.85\%}), showcasing SGG's potential to substantially improve alignment methods with favorable efficiency.

\subsection{Comparison Results with MLLMs}
\label{sec:comp_mllm}
We validate SGG's effectiveness in MLLMs, following LLaVA-v1.5 with a pretrained Vicuna-v1.5-7B~\cite{chiang2023vicuna}, pretrained 2$\times$MLP, and a pretrained CLIP~\cite{radford2021clip}, supervised fine-tuned for one epoch with a batch size of 64.
\textbf{(i) Full-Rank SFT:} AdamW, Adafactor, and LAMB are considered as baselines, with details of hyperparameters and settings provided in Table~\ref{tab:optimzer_hyper}, and display results of mainstream MLLM methods. The results in Table~\ref{table:comp_mllm_main} show that SGG boosts AdamW by \gbf{+1.0\%} on average. When paired with Adafactor, SGG could offer \gbf{+0.6\%} gains compared to baseline. Notably, SGG delivers an impressive \gbf{+2.4\%} improvement on VizWiz.
\textbf{(ii) PEFT and Quantization:} To rigorously evaluate SGG in resource-constrained scenarios, we conduct PEFT (LoRA) and 8-bit Quantization LoRA (Q-LoRA~\citep{dettmers2024qlora}) with rank $r=128$ and scaling factor $\alpha=256$.
Table~\ref{table:comp_mllm_main} shows that SGG achieves \textbf{65.1\%} average accuracy and yields \gbf{+2.2\%} gains over LoRA on VizWiz. Furthermore, SGG also enhances QLoRA (8-bit) SFT by \gbf{+0.6\%} on average.
All these results demonstrate SGG's versatility and effectiveness in boosting MLLM performance across SFT, PEFT, and quantized FT (Table~\ref{table:comp_mllm_full}).

\begin{figure}[t]
    \vspace{-0.5em}
    \centering
    \includegraphics[width=0.99\linewidth]{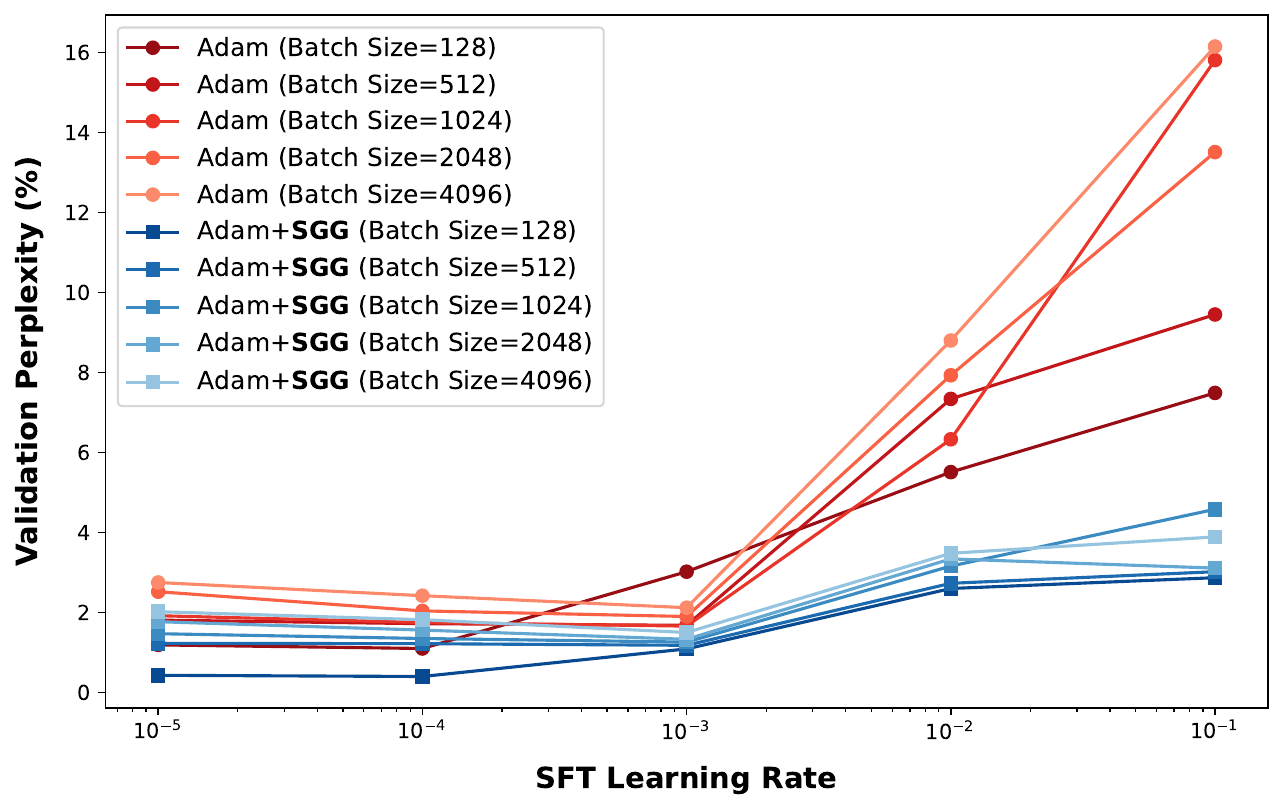}
    \vspace{-2.15em}
    \caption{\textbf{Learning Rate and Batch Size Scaling-up} with Qwen2.5-0.5B SFT on Alpaca. Validation loss$\downarrow$ vs SFT Learning Rate for \textcolor{fig3red}{\textbf{Adam}} and \textcolor{fig3blue}{\textbf{Adam+SGG}} across various batch sizes (128 to 4096). SGG offers consistent robustness over a wider range of hyper-parameters.
    }
    \label{fig:analysis_lr_vs_bs}
    \vspace{-0.75em}
\end{figure}

\subsection{Robustness to Learning Rate Scaling-up}
\label{sec:analysis}
Adam-like optimizers often struggle with the interplay between learning rate (LR) and batch size, leading to training instability (\textit{e.g.}, the surge phenomenon~\citep{li2024lr_scaling_law}) and meticulous tuning. In contrast, SGG shows exceptional robustness in this regard. During SFT on Alpaca~\citep{taori2023alpaca} with Adam (Figure~\ref{fig:analysis_lr_vs_bs}), SGG maintains stable validation loss across a wide spectrum of batch sizes ($128$ to $4096$) and learning rates, even under extreme conditions like batch sizes of $4096$ and LR of 0.1. This suggests that SGG \textit{effectively mitigates gradient outliers and dynamically adapts LRs}, ensuring reliable training across diverse configurations. Please refer to Appendix~\ref{app:analysis} for details.

\subsection{Ablation Studies}
\label{sec:ablation}
\begin{figure}[t]
    \centering
    \subfigtopskip=-0.5pt
    \subfigbottomskip=-0.5pt
    \subfigcapskip=-5.0pt
    \subfigure[Recluster Interval $T$]{\hspace{-0.25em}
    \label{fig:ablation(a)}\includegraphics[width=0.492\linewidth,trim= 0 0 0 0,clip]{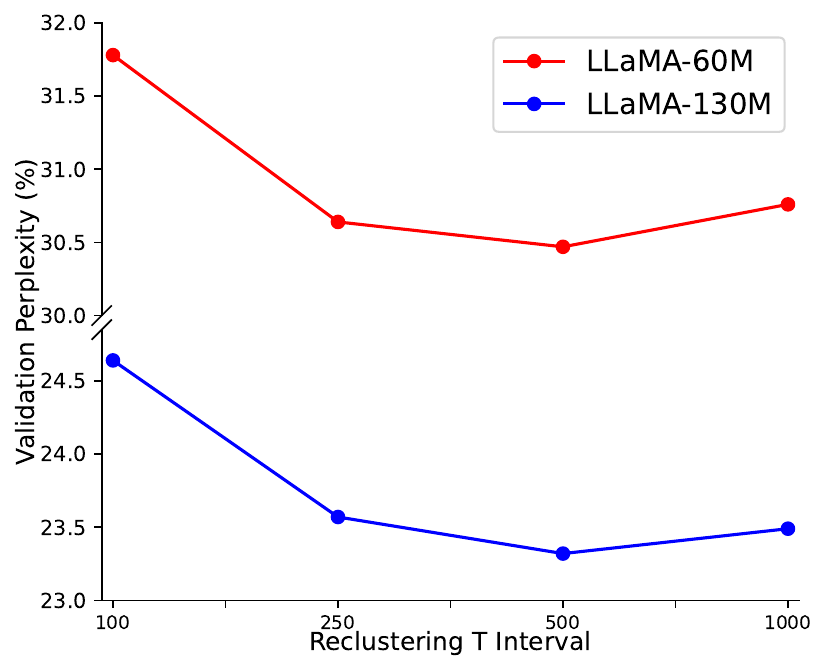}}
    \subfigure[Scaling Decay $\beta_3$]{\hspace{-0.25em}
    \label{fig:ablation(b)}\includegraphics[width=0.492\linewidth,trim= 0 0 0 0,clip]{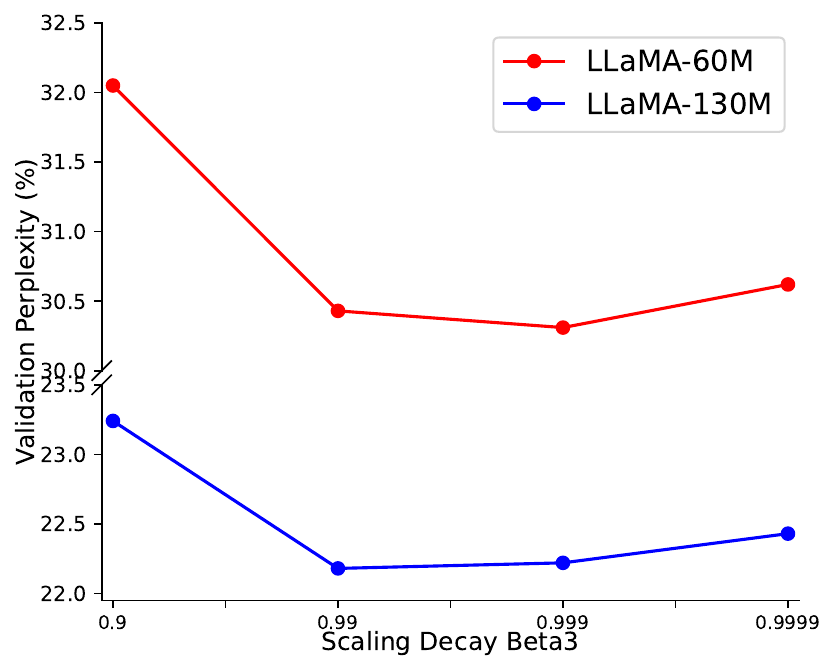}}
    \vspace{-0.85em}
    \caption{\textbf{Ablation of Hyperparameters} with LLaMA-60M and LLaMA-130M pre-training on C4. Validation Perplexity (PPL \%$\downarrow$: lower is better) vs. \textbf{(a)} Recluster Interval $T$ (\% total iterations) and \textbf{(b)} EMA Decay $\beta_3$. The results demonstrate that $T \approx 500$ and $\beta_3 = 0.99$ are the most favorable settings for SGG upon Adam.
    }
    \label{fig:ablation_hyper}
    \vspace{-1.0em}
\end{figure}

We analyze the three key hyper-parameters in SGG. \textbf{(i) Cluster number:} Table~\ref{tab:ablation} shows that $K$ can be easily set to \{2,3\} for diverse tasks according to the mini-batch K-means diagnostics. \textbf{(ii) Interval:} The interval $T$ can be set as 5\% of the total training iterations, \textit{e.g.}, $T=500$ for LLaMA-60M yields strong results, as shown in Figure~\ref{fig:ablation(a)}. \textbf{(iii) LR scaling decay:} Figure~\ref{fig:ablation(b)} demonstrates that SGG is insensitive to the precise value of scaling decay $\beta_3$, with $\beta_3 = 0.99$ proving a robust choice.

\begin{table}[htb]
    \centering
    \vspace{-0.25em}
    \caption{\textbf{Ablation studies} of the number of SGG clusters $K$ (task-relevant) across different tasks and models. ERR denotes the mini-batch K-means running errors.
    }
    \vspace{-0.60em}
    \setlength{\tabcolsep}{1.0mm}
\resizebox{1.0\linewidth}{!}{
    \begin{tabular}{l|c|cbbc}
    \toprule
Task                  & Model        & K=1 (Adam)  & K=2       & K=3       & K=4        \\ \hline
C4$\downarrow$        & LLaMA-60M    & \gray{34.1} & \bf{30.3} & 30.8      & \gray{ERR} \\
C4$\downarrow$        & LLaMA-130M   & \gray{25.1} & \bf{23.3} & 23.5      & \gray{ERR} \\
GLUE (MNLI)$\uparrow$ & RoBERT-Base  & \gray{87.2} & \bf{88.3} & 87.9      & \gray{ERR} \\
MLLM$\uparrow$        & LLaVA-1.5-7B & \gray{63.6} & 64.2      & \bf{64.5} & 64.3       \\
    \bottomrule
    \end{tabular}
    }
    \label{tab:ablation}
    \vspace{-0.5em}
\end{table}

\section{Related Work}
\label{sec:related}
\paragraph{Efficient Optimizers.}
Adaptive learning rate optimizers~\citep{iclr2019AdamW} are prevalent in training LLMs due to their balance of convergence speed and generalization. However, their effectiveness might diminish at scale because of the reliance on global gradient statistics, which overlook the inherent heterogeneity in LLMs~\citep{zhao2024deconstructing, zhang2025why_adam}. This heterogeneity, combined with the low-rank properties of LLMs, often leads to inefficient parameter updates and suboptimal convergence \citep{chen2024fira, zhao2024galore}. Traditional methods like Adam exhibit limitations in handling gradient dynamics under LLM low-rank constraints \citep{li2024lorap}, prompting the development of memory-efficient optimizers such as BAdam \citep{luo2025badam} and LISA \citep{pan2025lisa}. Techniques like Adam-mini \citep{zhang2024adam_mini} and APOLLO \citep{zhu2024apollo} further demonstrate that reduced learning rates or SGD-like memory footprints can achieve competitive performance. Nevertheless, challenges persist, particularly in scaling optimization for large models, as evidenced by the surge phenomenon in optimal learning rate and batch size scaling \citep{li2024lr_scaling_law}. Recent studies like SPAM \citep{huang2025spam} and CAME \citep{luo2023came} introduce momentum reset and confidence-guided strategies to stabilize training. SGG addresses these issues by grouping gradients and applying group-specific scaling, ensuring tailored learning rates.

\vspace{-0.5em}
\paragraph{Parameter-efficient Fine-tuning.}
Parameter-efficient Fine-tuning (PEFT) has become essential for adapting LLMs to downstream tasks efficiently. LoRA \citep{hu2021lora} as a foundational PEFT technique significantly reduces the computational costs by training only a small number of low-rank perturbation matrices added to the pre-trained weights. Recent extensions like DoRA variants \citep{liu2024dora, nasiri2025edora} further improve adaptation efficiency while maintaining performance. Despite their success, LoRA-based methods exhibit limitations: reliance on Dropout for regularization can be ineffective, especially in short training regimes \citep{kamalakara2022low_rank}. Suboptimal initialization can impede convergence in sparse data scenarios, and static scaling factors hinder adaptive tuning of learning rates  \citep{dettmers2023qlora}. While recent efforts like LoRA+ \citep{hayou2024lora+} and LoRA-XS \citep{balazy2024lora_xs} attempt to mitigate some of these issues, challenges persist, particularly in complex multi-modality perception tasks~\citep{ma2024vlora} and broader PEFT applications \citep{zhang2025peft_survey}. These limitations underscore the need for low-rank optimization that is migratable and can adjust learning rates with the low-rank property, which could be addressed by the gradient grouping-based learning rate scaling in SGG.
\section{Conclusion}
\label{sec:conclusion}
This paper presents SGG, an optimizer wrapper to address the challenges in LLM training. SGG clusters momentum vectors in each layer and computes cluster-specific scaling factors to modulate parameter-wise learning rates. Experiments demonstrate SGG's versatility and effectiveness with consistent performance gains and faster convergence when integrated with other optimizers and LoRA.

\section{Discussion and Limitations}
\paragraph{Border Impact.}
The expanding application of LLMs underscores the need for efficient and effective optimization. SGG offers a distinct approach: rather than relying on global approximation techniques (\textit{e.g.}, NMF) or architectural modifications (\textit{e.g.}, LoRA variants), SGG employs intra-layer gradient clustering and performs cluster-specific scaling to parameter-wise adaptive learning rates. This scheme allows seamless integration with mainstream optimizers and LoRA, yielding consistent performance gains across diverse (M)LLM applications with negligible additional cost.

\vspace{-0.25em}
\paragraph{Limitations and Discussion.}
While SGG shows great promise, its implementation shows avenues for future studies along this line: \textbf{(1) Grouping Strategies:} SGG's reliance on online clustering for dynamic grouping, though intuitive, represents only one specific choice. The adaptive learning rate scaling paradigm itself is flexible, which could include broader grouping designs, such as more precise online clustering, heuristic-based static partitioning, or even learned grouping functions, any of which might offer different performance-efficiency trade-offs for diverse scenarios and demands.
\textbf{(2) Computational Efficiency:}
While the CPU offloading in SGG mitigates the GPU burden, its online clustering still brings significant costs, which presents a huge concern in resource-constrained scenarios. Future work could focus on lightweight grouping methods or operations that could approximate grouping benefits without explicit clustering, thereby further enhancing the LLM optimizer's efficiency and applicability.
\textbf{(3) Evaluation Scope:} The validation in this work covers diverse benchmarks, yet extending the evaluation to a wider array of scenarios such as image generation \cite{ICLR2023MagViT2, cvpr2025MergeVQ}, multi-modalities \cite{iclr2022UnifiedIO, Xu2025Qwen25Omni}, architectures (\textit{e.g.}, vision backbones~\cite{cvpr2022convnext, iclr2024MogaNet} and Mixture-of-Experts~\cite{tkde2024MoESurvey}), and various data scales could provide deeper insights into SGG's generalization capabilities and potentially uncover new avenues for effective LLM training.

\section*{Acknowledgement}
This research is supported in part by the Early Career Scheme of the Research Grants Council (RGC) of the Hong Kong SAR under grant No. 26202321 and HKUST Startup Fund No. R9253.
This work was done when Juanxi Tian and Xin Jin interned at Westlake University and Peking University. The authors thank the Westlake University AI Station for supporting GPUs.



\bibliography{reference}

\clearpage
\renewcommand\thefigure{A\arabic{figure}}
\renewcommand\thetable{A\arabic{table}}
\setcounter{table}{0}
\setcounter{figure}{0}

\appendix


\section*{Appendix}

\section{Implementation Details}
\label{app:implement}
SGG is implemented in PyTorch and designed for seamless integration with mainstream adaptive optimizers such as Adam variants~\cite{iclr2015adam}. It requires no modifications to model architectures and only minimal additions to the optimization loop. SGG introduces a few key hyperparameters, which are empirically tuned to balance computational overhead with performance. It includes the number of clusters \(K \in \{2, 3\}\), the recluster interval $T\in [200, 1000]$ which is typically set to $1\sim5$\% of the total training iterations, and the scaling factor EMA decay \(\beta_3 = 0.99\). These hyperparameters are empirically tuned to balance computational efficiency and optimization performance. To minimize GPU memory footprint, especially for large-scale models, clustering indices, and scaling factors can be stored in CPU memory. This ensures that SGG remains scalable without imposing significant additional GPU memory demands.

The SGG wrapper operates in two main stages within each optimization step for layers designated for scaling: (i) gradient grouping by online clustering, and (ii) cluster-specific learning scaling for each parameter. For optimizers like \textbf{Adam}~\cite{iclr2015adam}, the momentum estimates \(m_l^t\) (which provide a smoothed representation of the gradients) are flattened and clustered instead. The clustering is performed using the \texttt{MiniBatchKMeans} algorithm from the \texttt{sklearn} library, which is efficient and suitable for large datasets. During clustering, the flattened gradients or momentum estimates are reshaped into a 2D array of shape \((N, 1)\), where \(N\) is the total number of elements in the gradient tensor. After clustering, each gradient element is assigned to a cluster, and the scaling factors \(\mathcal{S}_l\) are updated using an EMA of the median gradient magnitudes within each cluster. These scaling factors are then applied to the learning rates during the parameter update step, enabling adaptive and cluster-specific optimization. The immigration of SGG to other adaptive learning rate optimizers~\cite{icml2018adafactor, iclr2020lamb, liu2025muon, luo2023came} should be similar to this case.
The entire process is overall computationally efficient, with the extra clustering performed on the CPU and only the final scaling factors transferred to the GPU for parameter updates (the costs are nearly ignorable). Moreover, the proposed scaling operations will not be employed on the scalar and vector parameters like normalization layers and bias, as Muon, because these parameters do not have low-rank properties and are scale sensitive.

\begin{table}[t]
    \centering
    \caption{Hyperparameters of LLaMA models for pre-training and evaluation.}
    \vspace{-0.60em}
\resizebox{1.0\linewidth}{!}{
    \begin{tabular}{l|cccc}
    \toprule
\bf{Model}        & \bf{60M} & \bf{130M} & \bf{350M} & \bf{1B} \\ \hline
Embedding dim.    & 512      & 768       & 1024      & 2048    \\
Intermediate dim. & 1376     & 2048      & 2736      & 5461    \\
Heads             & 8        & 12        & 16        & 24      \\
Layers            & 8        & 12        & 24        & 32      \\ \hline
Steps             & 10K      & 20K       & 60K       & 100K    \\
Warmup            & 1K       & 2K        & 6K        & 10K     \\
Data Amount       & 1.3B     & 2.6B      & 7.8B      & 13.1B   \\
    \bottomrule
    \end{tabular}
    }
    \label{tab:c4_settings}
    \vspace{-0.5em}
\end{table}

\begin{table*}[t]
    \centering
    \vspace{-0.5em}
    \caption{\textbf{Full Comparison Results of LLaMA Pre-training on C4} using full-rank and memory-efficient training with the model sizes ranging from 60M to 1B. The validation perplexity (PPL$\downarrow$: lower is better) and GPU memory (Mem.)$\downarrow$ are reported, where only the weights and optimization states are considered. \textbf{Bold} and \gbf{green} types denote the best results and performance gains$\downarrow$ of SGG (\sethlcolor{lightblue}\hl{blue background}) over related baselines (\sethlcolor{gray90}\hl{gray background}).
    Note that $\dagger$ denotes results borrowed from previous papers, while others were reproduced by us.
    }
    \vspace{-0.70em}
\resizebox{0.87\linewidth}{!}{
    \begin{tabular}{lc|cccccccc}
    \toprule
\textbf{Method}              & \textbf{Venue}      & \multicolumn{2}{c}{\textbf{60M}} & \multicolumn{2}{c}{\textbf{130M}} & \multicolumn{2}{c}{\textbf{350M}} & \multicolumn{2}{c}{\textbf{1B}} \\
                             &                     & PPL            & Mem.            & PPL             & Mem.            & PPL             & Mem.            & PPL            & Mem.           \\ \hline
\grow Adam$^\dag$            & ICLR'15~            & 34.06          & 0.36G           & 25.08           & 0.76G           & 18.80           & 2.06G           & 15.56          & 7.80G          \\
NAdam                        & ICLR'18~            & 35.86          & 0.36G           & 28.88           & 0.76G           & 19.24           & 2.06G           & 15.78          & 7.80G          \\
RAdam                        & ICLR'20~            & 30.43          & 0.36G           & 25.17           & 0.76G           & 19.13           & 2.06G           & 15.65          & 7.80G          \\
LAMB                         & ICLR'20~            & 33.04          & 0.36G           & 24.37           & 0.77G           & 18.26           & 2.07G           & 15.84          & 7.81G          \\
Adan                         & TPAMI'23~           & 32.01          & 0.36G           & 23.14           & 0.77G           & 17.32           & 2.31G           & 14.70          & 15.78G         \\
Muon                         & arXiv'24~           & \bf{28.93}     & 0.36G           & \bf{22.34}      & 0.76G           & \bf{17.09}      & 2.06G           & \bf{14.52}     & 7.80G          \\
\brow Adam+\bf{SGG}          & \bf{Ours}           & 30.31          & 0.36G           & 22.18           & 0.76G           & 17.28           & 2.06G           & 14.30          & 7.80G          \\
\multicolumn{2}{l|}{\gray{$\Delta$ \textit{Gain}}} & \gbf{-3.75}    & \gray{+0.00}    & \gbf{-2.90}     & \gray{+0.00}    & \gbf{-1.52}     & \gray{+0.00}    & \gbf{-1.26}    & \gray{+0.00}   \\ \hline
Adafactor$^\dag$             & ICML'18~            & 32.57          & 0.24G           & 23.98           & 0.61G           & 17.74           & 1.53G           & 15.19          & 6.65G          \\
LION                         & arXiv'23~           & 50.89          & 0.34G           & 30.67           & 0.73G           & 21.28           & 1.98G           & 15.72          & 5.51G          \\
Low-Rank$^\dag$              & arXiv'22~           & 78.18          & 0.26G           & 45.51           & 0.54G           & 37.41           & 1.08G           & 34.53          & 3.57G          \\
Adam-mini$^\dag$             & ICLR'25~            & 34.10          & 0.23G           & 24.85           & 0.48G           & 19.05           & 1.32G           & 16.07          & 4.75G          \\
\grow CAME                   & ACL'23~             & 31.37          & 0.25G           & 23.38           & 0.62G           & 17.45           & 1.55G           & 14.68          & 6.70G          \\
\brow CAME+\bf{SGG}          & \bf{Ours}           & \bf{30.15}     & 0.25G           & 22.91           & 0.62G           & 17.09           & 1.55G           & 14.35          & 6.70G          \\
\multicolumn{2}{l|}{\gray{$\Delta$ \textit{Gain}}} & \gbf{-1.22}    & \gray{+0.00}    & \gbf{-0.46}     & \gray{+0.00}    & \gbf{-0.36}     & \gray{+0.00}    & \gbf{-0.33}    & \gray{+0.00}   \\
\grow APOLLO$^\dag$          & MLSys'25            & 31.55          & 0.24G           & 22.94           & 0.52G           & 16.85           & 1.22G           & 14.20          & 4.38G          \\
\brow APOLLO+\bf{SGG}        & \bf{Ours}           & 30.18          & 0.24G           & \bf{22.52}      & 0.52G           & \bf{16.54}      & 1.22G           & \bf{13.95}     & 4.38G          \\
\multicolumn{2}{l|}{\gray{$\Delta$ \textit{Gain}}} & \gbf{-1.37}    & \gray{+0.00}    & \gbf{-0.42}     & \gray{+0.00}    & \gbf{-0.31}     & \gray{+0.00}    & \gbf{-0.25}    & \gray{+0.00}   \\ \hline
\grow LoRA$^\dag$            & ICLR'22             & 34.99          & 0.36G           & 33.92           & 0.80G           & 25.58           & 1.76G           & 19.21          & 6.17G          \\
ReLoRA$^\dag$                & ICLR'23             & 37.04          & 0.36G           & 29.37           & 0.80G           & 29.08           & 1.76G           & 18.33          & 6.17G          \\
GaLore$^\dag$                & ICML'24             & 34.88          & 0.24G           & 25.36           & 0.52G           & 18.95           & 1.22G           & 15.64          & 4.38G          \\
GaLore+SPAM$^\dag$           & ICLR'25             & 32.39          & 0.24G           & 23.98           & 0.52G           & 18.28           & 1.22G           & \bf{14.73}     & 6.17G          \\
\brow LoRA+\bf{SGG}          & \bf{Ours}           & \bf{30.62}     & 0.36G           & \bf{23.62}      & 0.80G           & \bf{17.86}      & 1.76G           & \bf{14.73}     & 6.17G          \\
\multicolumn{2}{l|}{\gray{$\Delta$ \textit{Gain}}} & \gbf{-4.37}    & \gray{+0.00}    & \gbf{-10.30}    & \gray{+0.00}    & \gbf{-7.72}     & \gray{+0.00}    & \gbf{-4.48}    & \gray{+0.00}   \\ \hline
\multicolumn{2}{l|}{Training Tokens}               & \multicolumn{2}{c}{1.1B}         & \multicolumn{2}{c}{2.2B}          & \multicolumn{2}{c}{6.4B}          & \multicolumn{2}{c}{13.1B}       \\
    \bottomrule
    \end{tabular}
    }
    \label{tab:comp_c4_pt_full}
\end{table*}

\section{Experimental Setups and Results}
\label{app:comparison}

\subsection{LLM Pre-training on C4}
We conducted extensive pre-training experiments on LLaMA-based large language models using the C4 dataset. The C4 dataset, a meticulously cleaned and processed version of Common Crawl's web corpus, serves as a benchmark for pre-training language models and learning word representations. To closely replicate real-world pre-training conditions, we implemented a no-repetition training protocol over a substantial data volume, scaling our experiments across model sizes up to 7 billion parameters. We provide a comprehensive overview of the LLaMA architecture and the specific hyperparameters employed during pre-training (Table \ref{tab:c4_settings}). The hyperparameters are standardized across all model sizes, with a maximum sequence length of 256 tokens and a batch size of 131,000 tokens (\textit{i.e.}, the total batch size of 512 samples). For experiments of all optimizers, we implemented a learning rate warmup phase for the initial 10\% of the total training steps, followed by a cosine annealing schedule that gradually reduces the learning rate to 10\% of its initial value.

\begin{figure*}[t]
    \vspace{-0.25em}
    \centering
    \subfigtopskip=-0.5pt
    \subfigbottomskip=-0.5pt
    \subfigcapskip=-4.0pt

    \subfigure[Full-Rank Optimizers]{\hspace{-0.25em}
    \label{fig:app_c4(a)}\includegraphics[width=0.48\linewidth,trim= 0 0 0 0,clip]{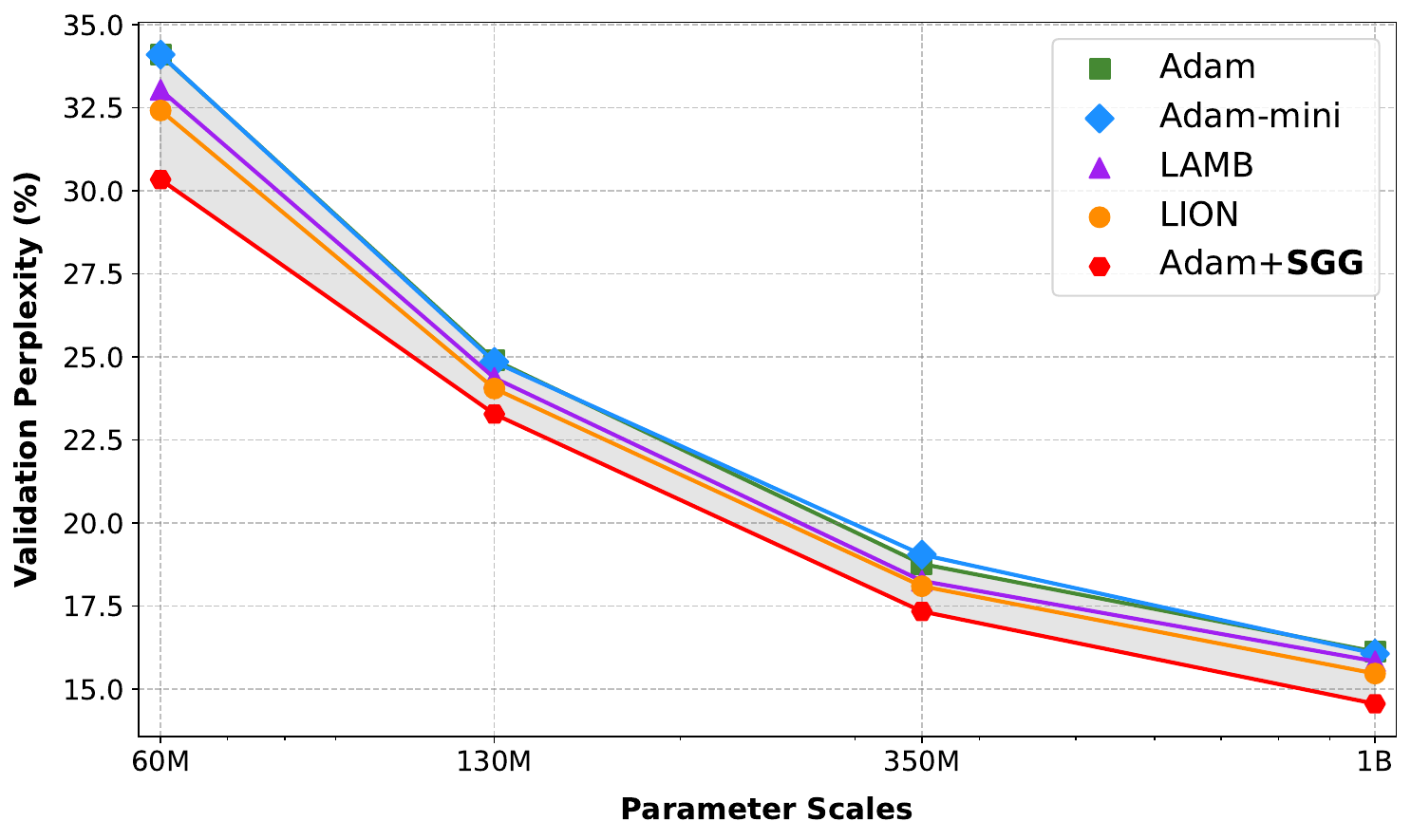}}
    \subfigure[Memory-efficient Optimizers]{\hspace{-0.25em}
    \label{fig:app_c4(b)}\includegraphics[width=0.48\linewidth,trim= 0 0 0 0,clip]{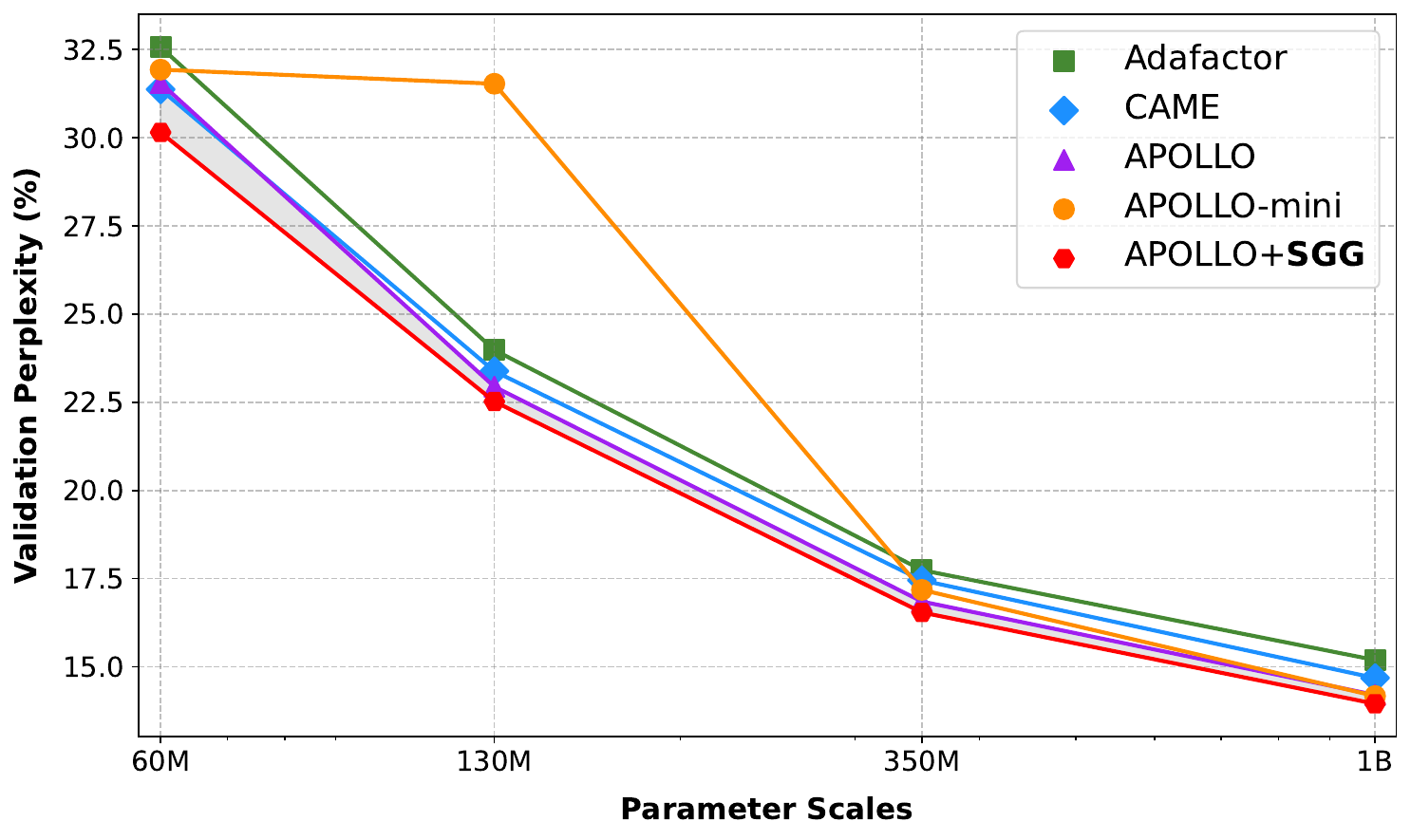}}
    \vspace{-0.5em}
    \caption{\textbf{Model Parameter Scaling-up Analysis} on C4 pre-training with different optimization algorithms.
    }
    \label{fig:app_c4_scaling}
    \vspace{-0.5em}
\end{figure*}

For each model size (ranging from 60 million to 1 billion parameters), we performed a systematic hyperparameter search to identify the optimal learning rate from the set \{1e-2, 5e-3, 1e-3, 5e-4, 1e-4\}, with selection criteria based on validation perplexity. Notably, SGG demonstrated remarkable robustness to hyperparameter variations, maintaining stable performance across different model sizes with a consistent learning rate. As shown in Table~\ref{tab:comp_c4_pt_full} and Figure~\ref{fig:app_c4_scaling}, we provided full benchmark results for the C4 pre-training experiments. We borrowed results of popular baselines from the previous studies, including Adam, Adam-mini~\cite{zhang2024adam_mini}, APOLLO~\cite{zhu2024apollo}, Low-Rank~\cite{kamalakara2022low_rank}, LoRA~\cite{hu2021lora}, ReLoRA~\cite{lialin2023relora}, GaLore~\cite{zhao2024galore}, SPAM~\cite{huang2025spam}, while reproducing more popular optimizers with the aforementioned experiments setups, including Adafactor~\cite{icml2018adafactor}, NAdam~\cite{iclr2018nadam}, RAdam~\cite{iclr2020radam}, LAMB~\cite{iclr2020lamb}, LION~\cite{nips2023lion}, CAME~\cite{luo2023came}, and Muon~\cite{liu2025muon}.

\subsection{LLM SFT on GLUE Benchmark}
\begin{table*}[ht]
\centering
\caption{Hyperparameters of fine-tuning RoBERTa base on GLUE benchmark.}
\vspace{-0.7em}
\label{tab:glue_settings}
\fontsize{10}{11}\selectfont 
\begin{tabular}{lcccccccc}
\toprule
 & MNLI & SST-2 & MRPC & CoLA & QNLI & QQP & RTE & STS-B \\
\midrule
Batch Size & 16 & 16 & 16 & 32 & 16 & 16 & 16 & 16 \\
\# Epochs & 30 & 30 & 30 & 30 & 30 & 30 & 30 & 30 \\
Learning Rate & 2e-05 & 1e-05 & 3e-05 & 3e-05 & 1e-05 & 1e-05 & 1e-05 & 2e-05 \\
Rank Config. & \multicolumn{8}{c}{Full} \\
Max Seq. Len. & \multicolumn{8}{c}{512} \\
\midrule
Batch Size & 16 & 16 & 16 & 32 & 16 & 16 & 16 & 16 \\
\# Epochs & 30 & 30 & 30 & 30 & 30 & 30 & 30 & 30 \\
Learning Rate & 2e-05 & 1e-05 & 3e-05 & 3e-05 & 1e-05 & 1e-05 & 1e-05 & 2e-05 \\
Rank Config. & \multicolumn{8}{c}{$r = 4$} \\
Max Seq. Len. & \multicolumn{8}{c}{512} \\
\midrule
Batch Size & 16 & 16 & 16 & 32 & 16 & 16 & 16 & 16 \\
\# Epochs & 30 & 30 & 30 & 30 & 30 & 30 & 30 & 30 \\
Learning Rate & 2e-05 & 2e-05 & 2e-05 & 3e-05 & 1e-05 & 2e-05 & 2e-05 & 3e-05 \\
Rank Config. & \multicolumn{8}{c}{$r = 8$} \\
Max Seq. Len. & \multicolumn{8}{c}{512} \\
\bottomrule
\end{tabular}
\end{table*}
The GLUE benchmark, a widely used evaluation framework for NLP tasks such as sentiment analysis, question answering, and textual entailment \cite{wang2018glue}, serves as a robust platform for assessing model performance. In this study, we fine-tuned the pre-trained RoBERTa-Base model on the GLUE benchmark using the Hugging Face implementation. The model was trained for 30 epochs with a batch size of 16 for all tasks except for CoLA, which utilized a batch size of 32. We meticulously tuned the learning rate and scale factor for the SGG optimization technique. Table~\ref{tab:glue_settings} details the hyperparameters employed for fine-tuning RoBERTa-Base with SGG.

The results, as presented in Table~\ref{table:comp_GLUE_full}, demonstrate the effectiveness of SGG in enhancing model performance across various GLUE sub-tasks. Notably, SGG consistently improves the top-1 accuracy when applied to different optimizers (AdamW and LAMB) with full-rank and low-rank settings. These enhancements underscore the advantage of SGG in stabilizing and accelerating the convergence of gradient-based optimization methods, particularly in low-rank settings where computational efficiency is crucial. The consistent performance gains across multiple tasks and optimizers highlight SGG's potential as a robust technique for fine-tuning large-scale language models, making it a valuable addition to the NLP toolkit.

\subsection{LLM PEFT with Commonsense Reasoning Tasks}
Following LLM-Adaptor~\citep{hu2023llmadapters}, we evaluate eight Commonsense Reasoning tasks with top-1 accuracy (\%) and GPU memory consumption, including BoolQ~\cite{clark2019boolq}, PIQA~\cite{bisk2020piqa}, SIQA~\cite{sap2019socialiqa}, HellaSwag~\cite{zellers2019hellaswag}, WinoGrande~\cite{sakaguchi2021winogrande}, ARC-Easy (ARC-E) and ARC-Challenge (ARC-C)~\cite{clark2018arc}, and OBQA~\cite{mihaylov2018obqa}.
As SFT setups in LLM-Adaptor, we combine the training datasets from all sub-tasks to fine-tune the pre-trained LLaMA-7B for 3 epochs using AdamW optimizer with a basic learning rate of 1e-4, a batch size of 32, and the rank $r=32$. Then, we evaluate each sub-task individually using its respective testing dataset.
Three classical PEFT baselines, Prefix-tuning (Prefix)~\citep{Li2021Prefix}, Series Adapter (Series)~\citep{Houlsby2019Series}, and Parallel Adapter (Parallel)~\citep{He2021Parallel}, and three popular PEFT methods, DoRA~\citep{liu2024dora}, GaLore~\citep{zhao2024galore}, and Fira~\citep{chen2024fira}, are compared in Table~\ref{tab:comp_cs_full}. Our SGG consistently improves eight sub-tasks over LoRA by \gbf{+2.9\%} without extra GPU memory, achieving competitive performances with well-designed PEFT methods with LoRA+SGG.

\begin{table*}[t]
    \centering
    \vspace{-0.5em}
    \caption{\textbf{Full Comparison Results of LLaMA PEFT} on eight commonsense reasoning datasets with the accuracy (\%$\uparrow$: higher is better) and the GPU memory$\downarrow$, where only the weights and optimization states are considered. ChatGPT results are obtained by Zero-shot CoT with gpt-3.5-turbo API. \textbf{Bold} and \gbf{green} types denote the best results and performance gains$\uparrow$ of SGG (\sethlcolor{lightblue}\hl{blue background}) compared to corresponding LoRA baselines (\sethlcolor{gray90}\hl{gray background}).
    }
    \vspace{-0.70em}
    \setlength{\tabcolsep}{0.8mm}
\resizebox{1.0\linewidth}{!}{
    \begin{tabular}{c|cc|cccccccc|c}
    \toprule
\textbf{Model} & \textbf{PEFT}                 & \textbf{Memory} & \textbf{BoolQ} & \textbf{PIQA} & \textbf{SIQA} & \textbf{HellaSwag} & \textbf{WinoGrande} & \textbf{Arc-E} & \textbf{Arc-C} & \textbf{OBQA} & \textbf{Average} \\ \hline
ChatGPT        & $-$                           & $-$             & \gray{73.1}    & \gray{85.4}   & \gray{68.5}   & \gray{78.5}        & \gray{66.1}         & \gray{89.8}    & \gray{79.9}    & \gray{74.8}   & \gray{77.0}      \\ \hline
               & Prefix                        & 0.05G           & 64.3           & 76.8          & 73.9          & 42.1               & 72.1                & 72.9           & 54.0           & 60.6          & 64.6             \\
               & Series                        & 0.42G           & 63.0           & 79.2          & 76.3          & 67.9               & 75.7                & 74.5           & 57.1           & 72.4          & 70.8             \\
               & Parallel                      & 1.49G           & 67.9           & 76.4          & 78.8          & 69.8               & 78.9                & 73.7           & 57.3           & 75.2          & 72.2             \\
\grow{}        & LoRA                          & 0.35G           & 68.9           & 80.7          & 77.4          & 78.1               & 78.8                & 77.8           & 61.3           & 74.8          & 74.7             \\
LLaMA-7B       & DoRA                          & 0.26G           & 69.7           & 83.4          & 78.6          & \textbf{87.2}      & 81.0                & 81.9           & \textbf{66.2}  & 79.2          & \textbf{78.4}    \\
               & GaLore                        & 0.26G           & 69.5           & 82.0          & 75.1          & 32.2               & 18.0                & 80.7           & 65.8           & 78.0          & 62.7             \\
               & Fira                          & 0.26G           & 69.4           & 82.6          & 78.0          & 76.8               & \textbf{81.2}       & \textbf{82.2}  & 64.4           & \textbf{80.8} & 76.9             \\
\brow{}        & LoRA+\bf{SGG}                 & 0.35G           & \textbf{70.3}  & \textbf{83.6} & \textbf{78.8} & 81.7               & 80.9                & 81.5           & 65.3           & 79.0          & 77.6             \\
               & \gray{$\Delta$ \textit{Gain}} & \gray{+0.00}    & \gbf{+1.4}     & \gbf{+2.9}    & \gbf{+1.4}    & \gbf{+3.6}         & \gbf{+2.1}          & \gbf{+3.7}     & \gbf{+4.0}     & \gbf{+4.2}    & \gbf{+2.9}       \\
    \bottomrule
    \end{tabular}
    }
    \label{tab:comp_cs_full}
    \vspace{-0.5em}
\end{table*}

\subsection{LLM RLHF with DPO}
In our experiments, we employed the Direct Preference Optimization (DPO) approach to fine-tune the Qwen2.5-0.5B model using the \textit{ultrafeedback\_binarized} dataset, which contains binary preference labels that facilitate the alignment of the model with human preferences~\cite{vonwerra2022trl}. The training process was conducted using both full-rank and LoRA strategies, with the latter being typically effective in reducing the number of trainable parameters while maintaining competitive performance. Hyperparameters include a learning rate of \(5.0 \times 10^{-7}\) for full-rank training and \(5.0 \times 10^{-6}\) for LoRA, with a single training epoch and a batch size of 2 per device. Gradient accumulation was set to 8 steps, and gradient checkpointing was enabled to optimize memory usage.

The optimization process utilized several optimizers, including SGD, AdamW, and LAMB, with and without the addition of the SGG (Stochastic Gradient with Gain) technique. As shown in Table~\ref{tab:comp_dpo}, the inclusion of SGG consistently improved the Top-1 accuracy across all optimizers. For instance, AdamW with SGG achieved a Top-1 accuracy of 71.85\% in full-rank training, representing a gain of \gbf{0.47\%} over the baseline AdamW. Similarly, in LoRA training, AdamW with SGG reached 72.02\%, a significant improvement of \gbf{1.80\%} compared to the baseline. These results underscore the advantage of SGG in enhancing the optimization process, particularly in scenarios where both computational efficiency and performance are critical.

The LoRA configuration used a rank (\(r\)) of 32 and alpha (\(\alpha\)) of 16, which provided a balance between model complexity and performance. The evaluation strategy was set in steps, with evaluations conducted every 50 steps, and logging was performed every 25 steps to monitor the training progress. The output directory was designated as \texttt{Qwen2-0.5B-DPO}, and the \texttt{no\_remove\_unused\_columns} flag was enabled to retain all columns in the dataset during training.

\subsection{MLLM SFT with LLaVA Variants}
To validate the generalization capability of the SGG-equipped optimizer, we also verify it on some variants of LLaVA~\cite{liu2024llava}. \emph{i.e.} LLaVA-v1.5-7b, LLaVA-LoRA, LLaVA-v1.3. And we choose some mainstream multi-modal LLMs at Table~\ref{table:comp_mllm_main}, \emph{e.g.} BLIP~\cite{li2022blip}, InstructBLIP~\cite{dai2023instructblip}, Qwen-VL~\cite{bai2023qwen}, Qwen-VL-Chat, mPLUG-Owl2~\cite{ye2024mplug}, and some variant of LLaVA, Tiny-LLaVA~\cite{zhou2024tinyllava}, MoE-LLaVA~\cite{lin2024moe-llava}, LLaVA-Phi~\cite{zhu2024llava-phi}, LLaVA-NeXT~\cite{liu2024llavanext}, LLaVA-MOD~\cite{shu2024llava-mod}, and LLaVA-KD-2B~\cite{cai2024llava-kd}.

\textbf{Setup and Settings:} Following the LLaVA-v1.5, we use a pre-trained Vicuna-v1.5-7B~\cite{chiang2023vicuna} as the language decoder. A pre-trained 2$\times$MLP is used as the connector to align the visual tokens to text tokens. The connector was trained by the \texttt{LCS-558K} datasets for one epoch. For the visual encoder, CLIP~\cite{radford2021clip} encodes and extracts the visual representation from the images. In our experiments, we validate three different optimizers: AdamW, Adafactor, and LAMB. We also reproduced results of popular optimizers like Muon, SOAP~\cite{Vyas2024SOAP}, and MARS~\cite{icml2025MARS} as the extension. The details of the optimizer hyperparameters and some training settings are shown in Table~\ref{tab:optimzer_hyper}.
\begin{table}[htb]
    \centering
    \caption{Details of the hyperparameters for the included optimizers and experiment settings.}
    \vspace{-0.7em}
    \setlength{\tabcolsep}{0.5mm}
\resizebox{1.0\linewidth}{!}{
    \begin{tabular}{c|ccc}
    \toprule
\bf{Method}    & \bf{AdamW}   & \bf{Adafactor} & \bf{LAMB}    \\ \hline
\multicolumn{4}{c}{\bf{Modules and datasets}}   \\ \hline
LLM            & \multicolumn{3}{c}{Vicuan-v1.5-7B}    \\
Vision encoder & \multicolumn{3}{c}{CLIP-L-336px}      \\
Connector      & \multicolumn{3}{c}{2$\times$MLP}      \\
Pretrain data  & \multicolumn{3}{c}{\texttt{LCS-558K}} \\  
SFT data  & \multicolumn{3}{c}{\texttt{llava-v1.5-mix665k}} \\ \hline
\multicolumn{4}{c}{\bf{Basic SFT settings}}   \\ \hline
Learning rate  & 2e$^{-5}$    & 2e$^{-5}$      & 2e$^{-5}$    \\
Batch size     & 64           & 64             & 64           \\
Betas          & (0.9, 0.999) & \xmarkg        & (0.9, 0.999) \\
Epsilon        & 1e$^{-8}$ & (1e$^{-30}$, 1e$^{-3}$) & 1e$^{-6}$ \\
Weight decay   & \xmarkg      & \xmarkg        & \xmarkg      \\
LR scheduler   & Cosine       & Cosine         & Cosine       \\
Warmup ratio   & 0.03         & 0.03           & 0.03         \\ 
Clip threshold & \xmarkg      & 1.0            & \xmarkg      \\ 
Clamp value    & \xmarkg      & \xmarkg        & 10           \\ 
Cluster number & 3            & 3              & 2            \\ 
Recluster interval  & 1,000   & 1,000          & 1,000        \\ 
Decay rate     & (0.95, 0.9)  & (0.95, 0.9)    & (0.95, 0.9)  \\ \hline
\multicolumn{4}{c}{\bf{Low-Rank hyperparameters}}   \\ \hline
LoRA ($r$=128, $\alpha$=256)      & \cmark       & \xmarkg        & \xmarkg      \\ 
8bit LoRA ($r$=128, $\alpha$=256)  & \cmark       & \xmarkg        & \xmarkg      \\

    \bottomrule
    \end{tabular}
    }
    \label{tab:optimzer_hyper}
\end{table}

\textbf{Supervised Fine-tuning:} We keep the visual encoder frozen and update the parameters of the connector and LLM for training. For the Full-Rank Supervised Fine-Tuning (SFT), the learning rate was set to 2e-5, the batch size was 64, and training one epoch on \texttt{llava-v1.5-mix665k} dataset. To further validate the effectiveness of SGG in the light parameters and low-bit quantization scenario, we conducted an experiment to train the Low-Rank (LoRA) and 8-bit Quantization LoRA (Q-LoRA~\cite{dettmers2024qlora}) SFT method. These methods have unique advantages in parameter efficiency and training speed. For the LoRA and Q-LoRA SFT, the rank $r$ of LoRA is 128, the learning rate scaling factor $\alpha$ is 256, the batch size set is 64, and training one epoch. These low-rank methods are based on the LLaVA-v1.5.

\textbf{Results:} Table~\ref{table:comp_mllm_main} and Table~\ref{table:comp_mllm_full} present the results of SGG on VQA and benchmark tasks. Table~\ref{table:comp_mllm_main} shows the results of seven representative tasks, while Table~\ref{table:comp_mllm_full} displays the full results of nine tasks. For the Full-Rank SFT, on the AdamW optimizer, SGG achieves 64.5 average performance on the 7 different tasks, which brings \gbf{+0.9\%} performance compared to the AdamW baseline. 
On the Adafactor, SGG could get extra \gbf{+0.5\%} performance compared to the vanilla Adafactor, especially on the VizWzi VQA task, SGG could bring \gbf{+2.4\%} capability. With LAMB+SGG, our performance can reach 52.7. For the LoRA SFT, our SGG could achieve 65.1 scores, and on the VizWiz task, it brings additional performance gains of \gbf{+2.2\%}. For the 8-bit experiments, Table~\ref{table:comp_mllm_main} shows that SGG with AdamW could also bring some performance.

\begin{table*}[t]
    \centering
    \caption{
    \textbf{Full Comparison Results with Mainstream MLLMs}. Compared with their counterparts, Top-1 accuracy (\%$\uparrow$: higher is better) is reported. AVG: The average of the nine benchmarks for comprehensive comparison, except for MME. $^\dag$: reproduced results using the official code. \gbf{Green types} denote the performance gains$\uparrow$ of SGG (\sethlcolor{lightblue}\hl{blue background}) over related baselines (\sethlcolor{gray90}\hl{gray background}). Most results are reported from LLaVA-KD~\cite{cai2024llava-kd}. 
    }
    \vspace{-0.75em}
    \renewcommand{\arraystretch}{1.2}
    \setlength\tabcolsep{0.2em}
    \resizebox{1.0\linewidth}{!}{
        \begin{tabular}{lcc|ccccc|cccccc}
        \toprule
        \multirow{2}{*}{\bf{Method}} & \multirow{2}{*}{\bf{LLM}} & \multirow{2}{*}{\bf{Optimizer}} & \multicolumn{5}{c}{\bf{Image Question Answering}} & \multicolumn{5}{c}{\bf{Benchmarks}} & \multirow{2}{*}{\bf{AVG}} \\
        \cline{4-8} \cline{9-13} 
                                &                 &                & VQAv2 & GQA  & VizWiz  & SciVQA$^I$ & TextVQA     & MME   & MMBench & MMBench$^{\text{CN}}$  & POPE & SEED$^I$ &                      \\\hline
        BLIP-2                  & Vicuna-13B      & AdamW          & 65.0  & 41.0 & 19.6    & 61.0  & 42.5        & $-$     & $-$     & $-$        & 85.3 & $-$   & $-$  \\
        InstructBLIP            & Vicuna-7B       & AdamW          & $-$   & 49.2 & 34.5    & 60.5  & 50.1        & $-$     & 36.0    & 23.7       & 79.8 & $-$   & $-$  \\
        Qwen-VL                 & Qwen-7B         & AdamW          & 78.8  & 59.3 & 35.2    & 67.1  & 63.8        & $-$     & 38.2    & 7.4        & $-$  & $-$   & $-$  \\
        Qwen-VL-Chat            & Qwen-7B         & AdamW          & 78.2  & 57.5 & 38.9    & 68.2  & 61.5        & $-$     & 60.6    & 56.7       & $-$  & $-$   & $-$  \\
        mPLUG-Owl2              & LLaMA2-7B       & AdamW          & 79.4  & 56.1 & 54.5    & 68.7  & 54.3        & $-$     & 66.5    & $-$        & 85.8 & $-$   & $-$  \\
        $\text{TinyLLaVA}^\dag$ & Qwen1.5-4B      & AdamW          & 79.9  & 63.4 & 46.3    & 72.9  & 59.0        & $-$     & 67.9    & 67.1       & 85.2 & $-$   & $-$  \\
        TinyLLaVA               & Phi2-2.7B       & AdamW          & 79.9  & 62.0 & $-$     & 69.1  & 59.1        & $-$     & 66.9    & $-$        & 86.4 & $-$   & $-$  \\
        Bunny                   & Phi2-2.7B       & AdamW          & 79.8  & 62.5 & 43.8    & 70.9  & 56.7        & $-$     & 68.6    & 37.2       & $-$  & $-$   & $-$  \\
        Imp-3B                  & Phi2-2.7B       & AdamW          & $-$   & 63.5 & 54.1    & 72.8  & 59.8        & $-$     & 72.9    & 46.7       & $-$  & $-$   & $-$  \\  
        MobileVLM               & MLLaMA-2.7B     & AdamW          & $-$   & 59.0 & $-$     & 61.0  & 47.5        & $-$     & 59.6    & $-$        & 84.9 & $-$   & $-$  \\
        MobileVLMv2             & MLLaMA-2.7B     & AdamW          & $-$   & 61.1 & $-$     & 70.0  & 57.5        & $-$     & 63.2    & $-$        & 84.7 & $-$   & $-$  \\
        MoE-LLaVA               & Phi2-2.7B       & AdamW          & 79.9  & 62.6 & $-$     & 70.3  & 57.0        & $-$     & 68.0    & $-$        & 85.7 & $-$   & $-$  \\
        LLaVA-Phi               & Phi2-2.7B       & AdamW          & 71.4  & $-$  & $-$     & 68.4  & 48.6        & $-$     & 59.8    & $-$        & 85.0 & $-$   & $-$  \\
        LLaVA-NeXT              & Vicuna-1.5-7B   & AdamW          & 81.8  & 64.2 & 57.6    & 70.1  & 64.9        & 1519.0  & 67.4    & 60.6       & 86.5 & 70.2  & 69.3 \\
        LLaVA-NeXT              & Vicuna-1.5-13B  & AdamW          & 82.8  & 65.4 & 60.5    & 73.6  & 67.1        & 1575.0  & 70.0    & 64.4       & 86.2 & 71.9  & 71.3  \\ 
        MiniCPM-V               & MiniCPM-2.4B    & AdamW          & $-$   & 51.5 & 50.5    & 74.4  & 56.6        & $-$     & 64.0    & 62.7       & 79.5 & $-$   & $-$  \\
        MiniCPMv2               & MiniCPM-2.4B    & AdamW          & $-$   & 52.1 & 60.2    & 76.3  & 73.2        & $-$     & 68.5    & 67.2       & 86.3 & $-$   & $-$  \\ 
        LLaVA-MOD               & Qwen1.5-1.8B    & AdamW          & $-$   & 58.7 & 39.2    & 68.0  & 58.5        & $-$     & 66.3    & 61.9       & 87.0 & $-$   & $-$  \\
        LLaVA-KD-2B             & Qwen1.5-1.8B    & AdamW          & 79.0  & 62.3 & 44.7    & 64.7  & 53.4        & $-$     & 64.0    & 63.7       & 86.3 & $-$   & $-$  \\ \hline
        \multicolumn{14}{l}{ \gray{\textit{LLaVA-v1.5/1.6 Full-Rank SFT}}}  \\
\grow   LLaVA-v1.5              & Vicuna-1.5-7B   & AdamW          & 78.5  & 62.0 & 50.0  & 66.8  & 58.2  & 1510.7  & 64.3    & 58.3  & 85.9 & 66.2  & 65.6  \\
\grow   LLaVA-v1.5              & Vicuna-1.5-7B   & Adafactor      & 79.0  & 62.7 & 48.2  & 70.7  & 57.1  & 1462.5  & 66.1    & 60.4  & 86.0 & 66.8  & 66.3   \\
\grow   LLaVA-v1.5              & Vicuna-1.5-7B   & LAMB           & 63.9  & 43.8 & 53.3  & 61.5  & 43.4  & 1090.9  & 43.2    & 41.8  & 81.2 & 50.4  & 53.6   \\ 
        LLaVA-v1.5              & Vicuna-1.5-7B   & Muon           & 79.3  & 62.6 & 50.3  & 69.1  & 57.7  & 1461.7  & 67.1    & 59.8  & 85.9 & 67.0  & 66.5   \\
        LLaVA-v1.5              & Vicuna-1.5-7B   & SOAP           & 79.4  & 62.5 & 47.8  & 69.7  & 57.9  & 1457.1  & 66.6    & 60.1  & 86.2 & 67.4  & 66.4   \\
        LLaVA-v1.5              & Vicuna-1.5-7B   & MARS           & 79.3  & 62.8 & 49.2  & 69.1  & 56.4  & 1451.1  & 66.7    & 59.4  & 86.1 & 67.5  & 66.3   \\ \hline
\brow   LLaVA-v1.5              & Vicuna-1.5-7B   & AdamW+\bf{SGG} & 79.1  & 62.4 & 50.2  & 69.8  & 57.4  & 1476.9  & 65.9    & 60.1  & 86.3 & 66.9  & 66.5 \\ 
        \multicolumn{3}{l|}{ \gray{$\Delta$ \textit{Gains compared to AdamW}}} & \gbf{+0.6}  & \gbf{+0.4} & \gbf{+0.2}  & \gbf{+2.0}  & \gbf{-0.8}  & \gbf{-33.8}  & \gbf{+1.6}    & \gbf{+1.8}  & \gbf{+0.4} & \gbf{+0.7}  & \gbf{+0.9}  \\
\brow   LLaVA-v1.5              & Vicuna-1.5-7B   & Adafactor+\bf{SGG} & 79.2 & 62.8 & 50.6  & 71.6  & 57.3  & 1477.2  & 66.3    & 60.8  & 86.0 & 67.3  & 66.8 \\
        \multicolumn{3}{l|}{ \gray{$\Delta$ \textit{Gains compared to Adafactor}}} & \gbf{+0.1}  & \gbf{+0.1} & \gbf{+2.4}  & \gbf{+0.9}  & \gbf{+0.2}  & \gbf{+14.7}  & \gbf{+0.2}    & \gbf{+0.4}  & \gbf{+0.0} & \gbf{+0.5}  & \gbf{+0.5}  \\ 
\brow   LLaVA-v1.5              & Vicuna-1.5-7B   & LAMB+\bf{SGG} & 64.3 & 44.0 & 53.3  & 61.8  & 43.5  & 1122.9  & 43.3    & 41.9  & 81.3 & 50.4  & 53.8 \\
        \multicolumn{3}{l|}{ \gray{$\Delta$ \textit{Gains compared to LAMB}}} & \gbf{+0.4}  & \gbf{+0.2} & \gbf{+0.0}  & \gbf{+0.3}  & \gbf{+0.1}  & \gbf{+32.0}  & \gbf{+0.1}    & \gbf{+0.1}  & \gbf{+0.1} & \gbf{+0.1}  & \gbf{+0.2}  \\ \hline
        \multicolumn{14}{l}{ \gray{\textit{LLaVA-v1.5 Low-Rank SFT (AdamW)}}}  \\
\grow   LLaVA-v1.5              & Vicuna-1.5-7B   & LoRA           & 79.1  & 63.0 & 47.8  & 68.4  & 58.2  & 1466.2  & 66.1    & 58.9  & 86.4 & 67.8  & 66.2  \\
\brow   LLaVA-v1.5              & Vicuna-1.5-7B   & LoRA+\bf{SGG}  & 79.1   & 63.4 & 51.0  & 70.1  & 58.6  & 1477.8  & 66.7    & 59.4  & 86.6 & 68.2  & 67.0   \\
        \multicolumn{3}{l|}{ \gray{$\Delta$ \textit{Gains compared to LoRA}}} & \gbf{+0.0}  & \gbf{+0.4} & \gbf{+2.2}  & \gbf{+1.5}  & \gbf{+0.4}  & \gbf{+11.6}  & \gbf{+0.6}    & \gbf{+0.5}  & \gbf{+0.2} & \gbf{+0.4}  & \gbf{+0.8} \\
    \bottomrule
    \end{tabular}
    \label{table:comp_mllm_full}
}
\end{table*}

\section{Empirical Analysis}
\label{app:analysis}

\subsection{Analysis of Gradient Clustering}
Figure~\ref{fig:sgg_clustering} illustrates the gradient clustering phenomenon observed during the pre-training of the LLaMA-1B model on the C4 dataset, focusing on gradients, adaptive learning rates, and gradient norms. LLMs exhibit unique gradient dynamics due to their massive scale, sparse activations, and hierarchical structure. SGG leverages these characteristics to improve optimization efficiency and convergence. Gradients in LLMs often follow a heavy-tailed distribution, with a small fraction of parameters contributing disproportionately to the overall gradient magnitude. SGG addresses this by flattening gradients into high-dimensional vectors and applying clustering algorithms (\textit{e.g.,} \( k \)-means) to group parameters with similar behaviors. This results in two distinct clusters: one for parameters with large gradients (associated with salient features or rare tokens) and another for those with smaller gradients (associated with frequent but less informative tokens). Adaptive learning rates are then computed separately for each cluster, ensuring stability for parameters with large gradients and faster convergence for those with smaller gradients. This contrasts with baselines that apply uniform learning rates, failing to account for the heavy-tailed gradient distributions typical of LLMs.

Figure~\ref{fig:c4(c)} depicts the layer-wise \( L_2 \)-gradient norm distributions across all layers of the LLaMA-1B model. Gradient norms vary significantly across layers due to the hierarchical nature of LLMs. Earlier layers (\textit{e.g.,} embedding and low-level transformer layers) exhibit smaller gradient norms, as they focus on general syntactic and semantic patterns. In contrast, deeper layers (\textit{e.g.,} higher-level transformer layers) tend to have larger gradient norms, as they model complex, context-dependent relationships. SGG captures these patterns by grouping parameters based on gradient norms and applying layer-wise learning rate scaling. This ensures that earlier layers receive larger updates for faster learning of general patterns, while deeper layers receive smaller updates to maintain stability and prevent overfitting. Baseline methods, which lack such adaptive scaling, often struggle to optimize all layers simultaneously, leading to suboptimal convergence and poor generalization.

The clustering of gradients, adaptive learning rates, and gradient norms in LLMs are deeply interconnected phenomena. The heavy-tailed gradient distribution directly influences adaptive learning rates, as parameters with large gradients are assigned smaller learning rates to prevent instability. This, in turn, affects gradient norms, as learning rate scaling impacts the magnitude of parameter updates. SGG's ability to capture these relationships and adaptively scale learning rates based on gradient clustering and norm distributions leads to more stable and efficient optimization compared to baseline methods. Furthermore, the hierarchical structure of LLMs introduces additional complexity, as different layers exhibit distinct gradient behaviors. SGG addresses this by leveraging layer-wise clustering and scaling, ensuring each layer is optimized according to its specific role. This is particularly critical for LLMs, where the interplay between low-level and high-level features is essential for capturing the nuances of natural language. By preserving the inherent structure of the optimization landscape, SGG not only improves convergence but also enhances the model's ability to generalize to unseen data.

\subsection{Analysis of Learning Rate Scaling}
We analyze the impact of learning rate scaling on the validation perplexity of the Qwen2.5-0.5B model fine-tuned on the Alpaca dataset. The experiments were conducted with varying batch sizes \{128, 512, 1024, 2048, 4096\} and learning rates \{1e-1, 1e-2, 1e-3, 1e-4, 1e-5\}, using both the Adam optimizer and Adam with SGG. The model was trained for 3 epochs with LoRA (rank=8) and followed the official settings of the Alpaca framework. The results, as depicted in Figure~\ref{fig:analysis_lr_vs_bs}, demonstrate several key trends. First, as the batch size increases, the validation perplexity generally decreases, indicating that larger batch sizes contribute to more stable and efficient training. This effect is particularly pronounced when SGG is applied, suggesting that SGG enhances the model's ability to generalize even under extreme batch size settings. Second, lower learning rates (\textit{e.g.,} 1e-4, 1e-5) consistently yield better performance, especially when combined with larger batch sizes, highlighting the importance of balancing these hyperparameters. 
Notably, SGG provides robust performance gains across all configurations, significantly reducing validation perplexity compared to standard Adam optimization. This improvement is attributed to SGG's ability to guide the optimization process more effectively, particularly in scenarios with large batch sizes and varying learning rates. Overall, the results underscore the effectiveness of SGG in enhancing model performance, even in challenging training conditions, and emphasize the critical role of hyperparameter tuning in achieving optimal results.

\end{document}